\documentclass[letterpaper]{article} 
\usepackage[submission]{aaai25}  
\usepackage{times}  
\usepackage{helvet}  
\usepackage{courier}  
\usepackage[hyphens]{url}  
\usepackage{graphicx} 
\usepackage{natbib}  
\usepackage{caption} 
\usepackage{algorithm}

\usepackage{xcolor}
\usepackage{babel}
\usepackage{array}
\usepackage{booktabs}
\usepackage{textcomp}
\usepackage{mathtools}
\usepackage{multirow}
\usepackage{amsmath}
\usepackage{amssymb}
\usepackage{subfig}
\usepackage{algpseudocode}
\usepackage{newfloat}
\usepackage{listings}
\DeclareCaptionStyle{ruled}{labelfont=normalfont,labelsep=colon,strut=off} 
\floatstyle{ruled}
\newfloat{listing}{tb}{lst}{}
\floatname{listing}{Listing}
\title{Luminance Component Analysis for Exposure Correction}

\author {
    Jingchao Peng\textsuperscript{\rm 1}\textsuperscript{\rm 2}, 
    Thomas Bashford-Rogers\textsuperscript{\rm 1}, 
    Jingkun Chen\textsuperscript{\rm 3}, 
    Haitao Zhao\textsuperscript{\rm 2}\thanks{Corresponding author.}, 
    Zhengwei Hu\textsuperscript{\rm 1}, 
    Kurt Debattista\textsuperscript{\rm 1}
}
\affiliations {
    \textsuperscript{\rm 1}WMG, University of Warwick\\
    \textsuperscript{\rm 2}School of Information Science and Technology, East China University of Science and Technology\\
    \textsuperscript{\rm 3}Institute of Biomedical Engineering, Department of Engineering Science, University of Oxford
}

\usepackage{bibentry}

\begin{document}

\maketitle

\begin{abstract}
Exposure correction methods aim to adjust the luminance while maintaining other luminance-unrelated information.
However, current exposure correction methods have difficulty in fully separating luminance-related and luminance-unrelated components, leading to distortions in color, loss of detail, and requiring extra restoration procedures.
Inspired by principal component analysis (PCA), this paper proposes an exposure correction method called luminance component analysis (LCA).
LCA applies the orthogonal constraint to a U-Net structure to decouple luminance-related and luminance-unrelated features.
With decoupled luminance-related features, LCA adjusts only the luminance-related components while keeping the luminance-unrelated components unchanged.
To optimize the orthogonal constraint problem, LCA employs a geometric optimization algorithm, which converts the constrained problem in Euclidean space to an unconstrained problem in orthogonal Stiefel manifolds.
Extensive experiments show that LCA can decouple the luminance feature from the RGB color space.
Moreover, LCA achieves the best PSNR (21.33) and SSIM (0.88) in the exposure correction dataset with 28.72 FPS.
\end{abstract}

%

\section{Introduction}
As computer vision (CV) continues to evolve, image quality is a critical factor in the success of various applications. 
The majority of CV tasks depend on images; however, variations in lighting conditions often compromise image quality \cite{DecomNet, discrete_cosine, Fourier}. 
Consequently, correcting exposure values in these images becomes essential. 
Exposure correction not only improves the visual appeal of images but also significantly enhances the performance of high-level vision tasks.
This process involves adjusting the exposure levels to achieve a visually clear image while preserving luminance-unrelated features \cite{baseline, detail_useless}.

\begin{figure}[!t]
\vspace{-0.3cm}
\includegraphics[width=1\columnwidth]{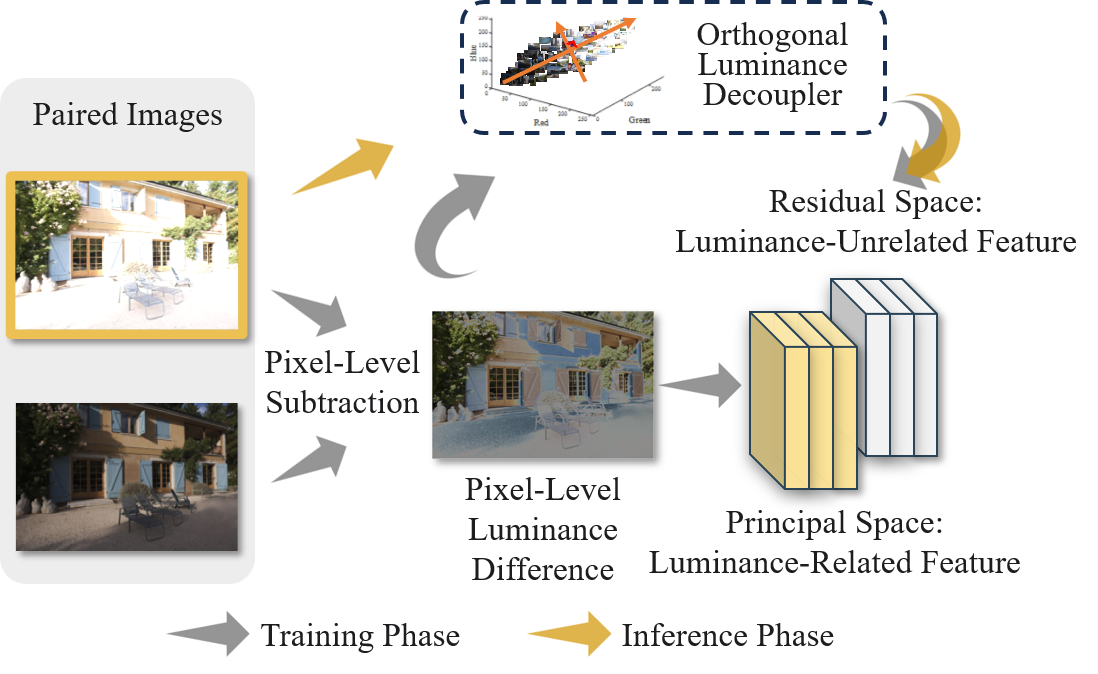}
\caption{The main idea of LCA. 
In the training phase the luminance information is first extracted from two identical images with different exposure values, then the luminance-related and luminance-unrelated components are decoupled by orthogonal decomposition.
During the inference phase, the well-trained decomposition modules are directly applied to the input image without the ground truth.
}
\label{fig:motivation-of-the-pro}
\vspace{-0.2cm}
\end{figure}

In color science, it is common to separate color into luminance-related and luminance-unrelated components \cite{color_space}.
This approach is exemplified by JPEG and other compression methods, which leverage the fact that the human visual system (HVS) is generally more sensitive to luminance than to color \cite{JPEG, color_compression_luminance}.
The underlying reason for this difference is that the HVS consists of rods, which handle luminance, and cones, which manage color perception \cite{HVS1, HVS2, HVS3}.
This indicates that luminance and color information are processed independently.
As a result, many color spaces adopt this separation to enhance color representation and facilitate more effective image processing.

Inspired by HVS and color space representation, many exposure correction methods decouple luminance-related and luminance-unrelated components and correct the exposure in the luminance-related components.
For example, transferring images from RGB color space into existing luminance-separated color spaces is a common method, such as when using cylindrical-coordinate representations (including HSV, HSI, and HSL) \cite{HSV} or YUV color space (including YUV, YIQ, and YCbCr) \cite{YCbCr}, in which there is one component representing the luminance.
Retinex-theory-based methods decompose images into illumination and reflectance components, then separately enhance the lightness and structure information \cite{Retinex1, Retinex2, Retinex3}.
Methods based on frequency domain decomposition decompose the original images into the frequency domain, where phase and amplitude components relate to structures and lightness, respectively \cite{Fourier,discrete_cosine}. 

Decoupling luminance-related and unrelated components is a non-linear and complex process.
Luminance-related components primarily reflect photometric intensity, influenced by light wavelength, amplitude, contrast, and context-based perception \cite{luminance_defination}.
They also affect saturation, with properly exposed images having higher saturation than over- or under-exposed ones.
In contrast, luminance-unrelated components encompass chromaticity, which includes color purity and dominant wavelength independent of brightness.
They also involve texture and edge information, maintaining stable visual perception despite changes in lighting and contrast.
This stability ensures consistent semantics information across various environments.
The current decomposition techniques are linear and non-orthogonal, such as having the luminance component calculated by $0.26\times R+0.50\times G+0.10\times B+16$ in YCrCb and $\textrm{max}(R, G, B)$ in HSV.
These straightforward decomposition techniques could entail that after decomposition, there are still close connections between the luminance-related and luminance-unrelated components.

The exposure correction performance suffers from incomplete decoupling.
According to experiments, Feng et al. \cite{detail_useless} and Huang et al. \cite{Fourier} showed that the luminance-unrelated components do not promote the performance of exposure correction.
Furthermore, Kumar et al. argued that any alterations to the luminance-unrelated color information (specifically, the Hue channel in the HSV color space) could result in color distortion and the loss of the original image color \cite{HSV}.
Due to the above reasons, most existing deep-learning-based (DL-based) methods need other subnetworks to handle the consequences of changing luminance-unrelated components, such as noise, color distortion, and the loss of structure and details \cite{HoLoCo, Fourier, LA-Net, detail_useless, RUAS, You_Only_Need_90K_P}.
The additional subnetworks may affect the inference speed and increase the difficulty of multi-objective optimization.
Therefore, it is best to preserve the luminance-unrelated information while only adjusting the luminance-related components.

Motivated by the above analysis, we propose a novel exposure correction method called luminance component analysis (LCA) in this paper.
LCA is based on the assumption that the difference between two identical images with different exposure values is the luminance information.
LCA borrows ideas from PCA and applies orthogonal decomposition to decouple the luminance-related and luminance-unrelated components, as shown in Fig. \ref{fig:motivation-of-the-pro}.
Instead of a traditional method, LCA is a DL-based method, which employs a U-Net structure as the baseline model. It recursively uses orthogonal luminance decouplers (OLD) to separate the luminance information into the luminance-related components.
The luminance-related and luminance-unrelated components can be decoupled, and the luminance-unrelated components remain unchanged while only the luminance components are adjusted for exposure correction.

To optimize the orthogonal constrained problem, we adopt a geometric optimization algorithm.
Specifically, we convert the constrained problem in Euclidean space to an unconstrained problem in the orthogonal Stiefel manifold.
Then, we compute the Riemannian gradient, the equivalent steepest-ascent direction in the Stiefel manifold.
Finally, a retraction transform guarantees each iteration point satisfies the orthogonal constraint.

In summary, our contributions are: 
\begin{enumerate}
\item Via the orthogonal constraint applied in deep learning, LCA can decouple the luminance-related and luminance-unrelated components.
\item With the decoupled luminance-related and luminance-unrelated components, LCA only adjusts the luminance-related components while keeping the luminance-unrelated components, which can minimize color distortion and the loss of structure and details.
\item LCA adopts a geometric optimization algorithm to convert the constrained problem to an unconstrained problem.
Most importantly, we theoretically prove its convergence.
\end{enumerate}

\section{Related Works}

\noindent \textbf{Exposure Correction.}

Traditional exposure correction methods utilize histogram equalization \cite{histogram3}, Retinex theory \cite{Retinex1,Retinex2,Retinex3}, gamma correction \cite{gamma_correction}, color correction matrix \cite{color_correction_matrix}, and other image processing techniques \cite{image_processing_techniques1,image_processing_techniques2} to correct poorly exposed images.
These methods are straightforward and fast.
However, when dealing with images with uneven illumination, they do not always perform effectively as they cannot automatically balance all appearance factors such as contrast, brightness, and saturation \cite{traditional_methods_not_work}.

Deep-learning-based methods decompose the image into a luminance space and another luminance-unrelated space, and they separately adopt subnetworks to handle the two spaces individually.
For example, Kumar et al. \cite{HSV} transformed RGB images into the HSV color space and fused multi-exposure images in the intensity channel, which is the brightness of the color representing the luminance information.
HoLoCo \cite{HoLoCo}, DecomNet \cite{DecomNet}, and RUAS \cite{RUAS} utilized the Retinex decomposition to decompose the luminance information.
LA-Net \cite{LA-Net} decomposed the luminance information through total-variation loss.
Furthermore, Laplacian pyramid decomposition \cite{detail_useless}, Fourier transform \cite{Fourier}, and discrete cosine transform \cite{discrete_cosine} were also employed to extract luminance. 

These methods, however, cannot minimize the relation between luminance-related and unrelated features, leading to additional processing \cite{detail_useless, RUAS} or under-performance of exposure correction \cite{HSV}. 

\noindent \textbf{Orthogonal Constraint.}
The orthogonal constraint can reduce correlated and redundant information \cite{orthogonal_constraint1}, prevent the potential risk of overfitting \cite{orthogonal_constraint2}, and remove the correlation between samples \cite{orthogonal_constraint3}.
To solve the orthogonal constrained problem in deep learning models, the constrained problem can be transformed into an unconstrained form by Lagrange multipliers or a barrier penalty function.
However, these methods do not fully utilize underlying orthogonal structures but only treat the problem as a \textquotedblleft black box\textquotedblright{}, solving it through algebraic manipulation \cite{manifold_survey}.
This may increase the optimization difficulty and introduce a new problem of balancing multi-objectives.

Manifold-based methods offer a solution by converting constrained problems in Euclidean space to unconstrained problems in Stiefel manifolds.
For instance, Peng et al. proposed a geometric conjugate gradient algorithm and trained a two-layer orthogonal autoencoder for fault detection \cite{SCA}.
Li et al. deployed the Cayley transform to project gradient and momentum vectors onto the tangent space and successfully trained the RNNs with the orthogonal constraint \cite{stiefel_manifold_SGD}.
Wang et al. applied original constraint into contrastive learning for effective image dehazing \cite{ODCR}.
These techniques highlight the potential of Stiefel manifold-based methods in deep learning.
\emph{However, these methods adjust the search direction rather than using the negative gradient direction, leaving their convergence unproven}.

\section{Methodology}

This section first presents the fundamental model and its solution, followed by a demonstration of its effectiveness using a toy problem. 
Subsequently, a proof-of-concept model applies this approach to the exposure correction task.

\subsection{Orthogonal Luminance Decoupler}

\begin{figure}[!t]
\begin{centering}
\includegraphics[width=0.7\columnwidth]{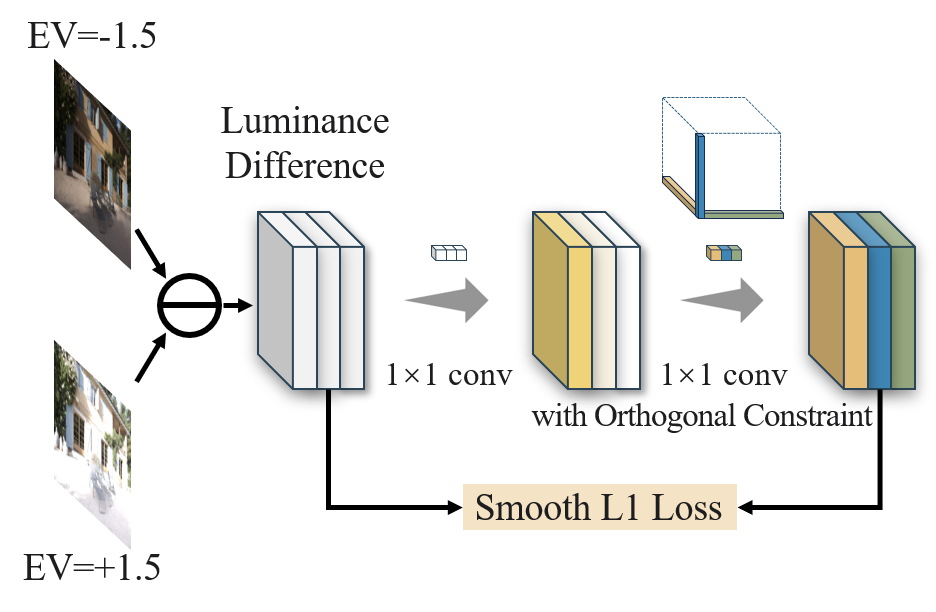}
\par\end{centering}
\caption{\label{fig:Toy-problem-str}The structure of OLD.}
\vspace{-0.5cm}
\end{figure}

The orthogonal luminance decoupler (OLD) employs an encoder-decoder structure with the orthogonal constraint to project image features into an orthogonal space to reduce the relevance between luminance-related and luminance-unrelated features.
The structure of OLD can be shown in Fig. \ref{fig:Toy-problem-str}.
First, the input and ground truth images are subtracted to obtain the luminance difference.
Then, OLD minimizes the reconstruction error of the luminance difference between the input ($In$) and ground truth ($GT$), such that:
\begin{equation}
\resizebox{0.42\textwidth}{!}{
\ensuremath{
\begin{array}{c}
\ensuremath{[W,\tilde{W}]=\underset{W,\tilde{W}}{\textrm{\ensuremath{\arg\ \min}}}\left\Vert (GT-In)-\left(\tilde{W}^{T}\left(W(GT-In)\right)\right)\right\Vert ^{2}}\\
s.t.\ensuremath{\tilde{W}^{T}\tilde{W}=I_{c_{in}}},
\end{array}}\label{eq:old-problem}}
\end{equation}
where $W\in\mathbb{R}^{c_{out}\times c_{in}}$ and $\tilde{W}\in\mathbb{R}^{c_{out}\times c_{in}}$ are the parameter matrix of the encoder $En(\cdot)$ and the decoder $De(\cdot)$, respectively;
$c_{in}$ and $c_{out}$ are the channel sizes of input and output of OLD, respectively.

The input and ground truth are identical images with varying exposure values; their luminance-unrelated information, such as the original hue and structure, should be the same.
Therefore, subtracting the input and ground truth can obtain the luminance difference.
With the orthogonal constraint, OLD can project the raw feature space into two subspaces, the principal and residual subspaces, the same as principal component analysis (PCA).
The principal subspace contains the majority of the luminance difference, which is made up of luminance-related components. 
In contrast, the residual subspace represents the luminance-unrelated information.\\

\begin{figure}[!b]
\vspace{-0.3cm}
\begin{centering}
\includegraphics[width=1.0\columnwidth]{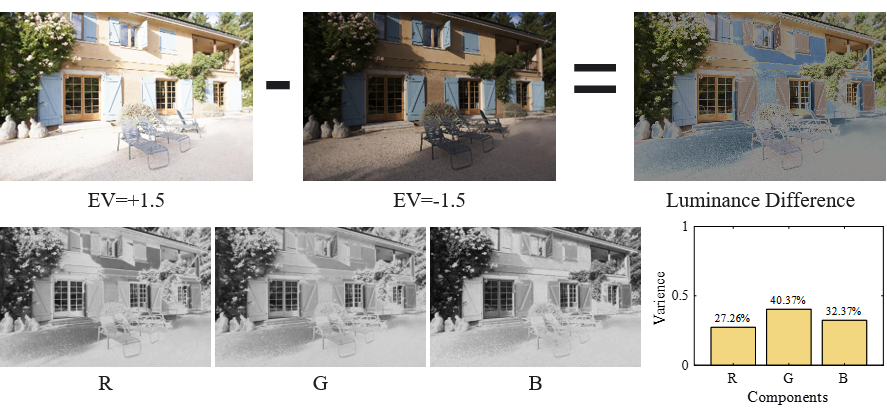}
\par\end{centering}
\vspace{-0.3cm}
\caption{\label{fig:Toy-problem-discription}Toy problem.
Subtracting two identical images with different exposure reveals luminance differences (representing luminance information).
Luminance information is coupled in the original RGB space, as indicated by high R, G, and B component variance.
}
\vspace{-0.5cm}
\end{figure}

\begin{figure*}[!t]
\makebox[\textwidth][c]{\includegraphics[width=1.05\textwidth]{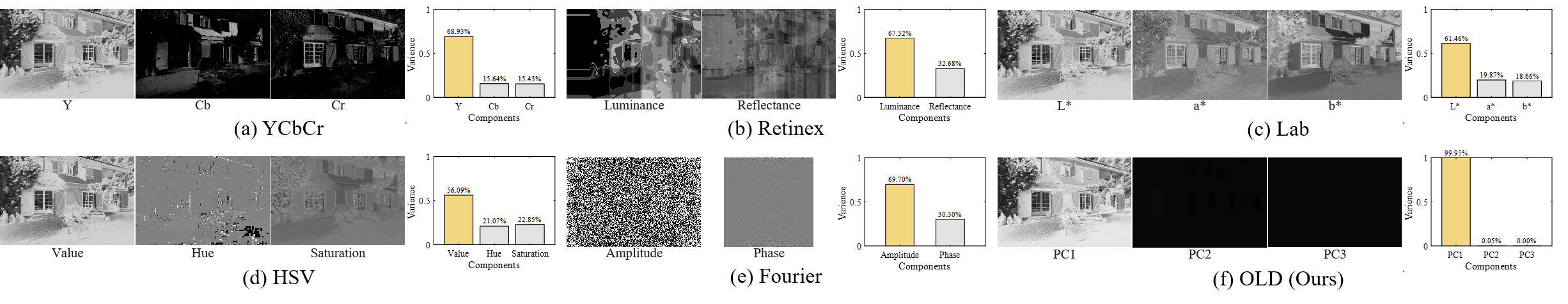}}
\caption{\label{fig:The-comparison-of-de}Decoupling the luminance-related feature by different methods.
Higher variance indicates the more luminance information the corresponding component contains.}
\vspace{-0.3cm}
\end{figure*}
      
\noindent \textbf{Toy Problem.}
To demonstrate the decoupling capability of the proposed OLD, we use an image pair with different exposure levels from the exposure correction dataset \cite{laplacian} as a toy problem, illustrated in Fig. \ref{fig:Toy-problem-discription}.
We aim to decouple the luminance-related components.
First, we subtract the two images, which, being identical except for exposure, should result in luminance-related components, as shown in the first line of Fig. \ref{fig:Toy-problem-discription}.
The subtraction reveals luminance information across all RGB channels, as shown in the R, G, and B channels in the second line of Fig. \ref{fig:Toy-problem-discription}, indicating that luminance is coupled with these channels.

We also calculate the variance of each channel.
With the same meaning as in PCA, higher variance represents greater data variability and information retention. 
Since the images differ only in exposure, the subtraction results should solely contain luminance-related information, with higher variance indicating more retained luminance-related information. 
Fig. \ref{fig:Toy-problem-discription} shows that the R, G, and B channels are all related to luminance in the raw RGB space, implying coupling across these channels.

We apply various methods (YCbCr, Retinex, Lab, HSV, and Fourier) to validate if the luminance-related components can be decoupled.
These methods map the luminance difference into different spaces. 
Then, the variance of each component is analyzed, as shown in Fig. \ref{fig:The-comparison-of-de}.
although existing decomposition methods can separate luminance information to some extent, there remain correlations between related and unrelated components, as indicated by the large variance in the luminance-unrelated components. 
The sum of variances of their luminance-unrelated components is greater than 30.30\%.
However, in our OLD, the sum of variances of the luminance-unrelated components is only 0.05\%, as shown on Fig. \ref{fig:The-comparison-of-de} (f).
This demonstrates that the proposed OLD can effectively separate the luminance-related and unrelated information with minimal correlation.

\subsection{Geometric Optimisation on Stiefel Manifolds\label{subsec:3.3}}
Optimizing the constrained problem  Eq. \eqref{eq:old-problem} is more difficult than an unconstrained problem.
One way is converting them into unconstrained forms by the Lagrange multiplier method or a barrier penalty function.
However, these methods treat the problem as a \textquotedblleft black box\textquotedblright{} while ignoring the underlying orthogonal relationship.

\begin{figure}[!t]
\begin{centering}
\includegraphics[width=1.0\columnwidth]{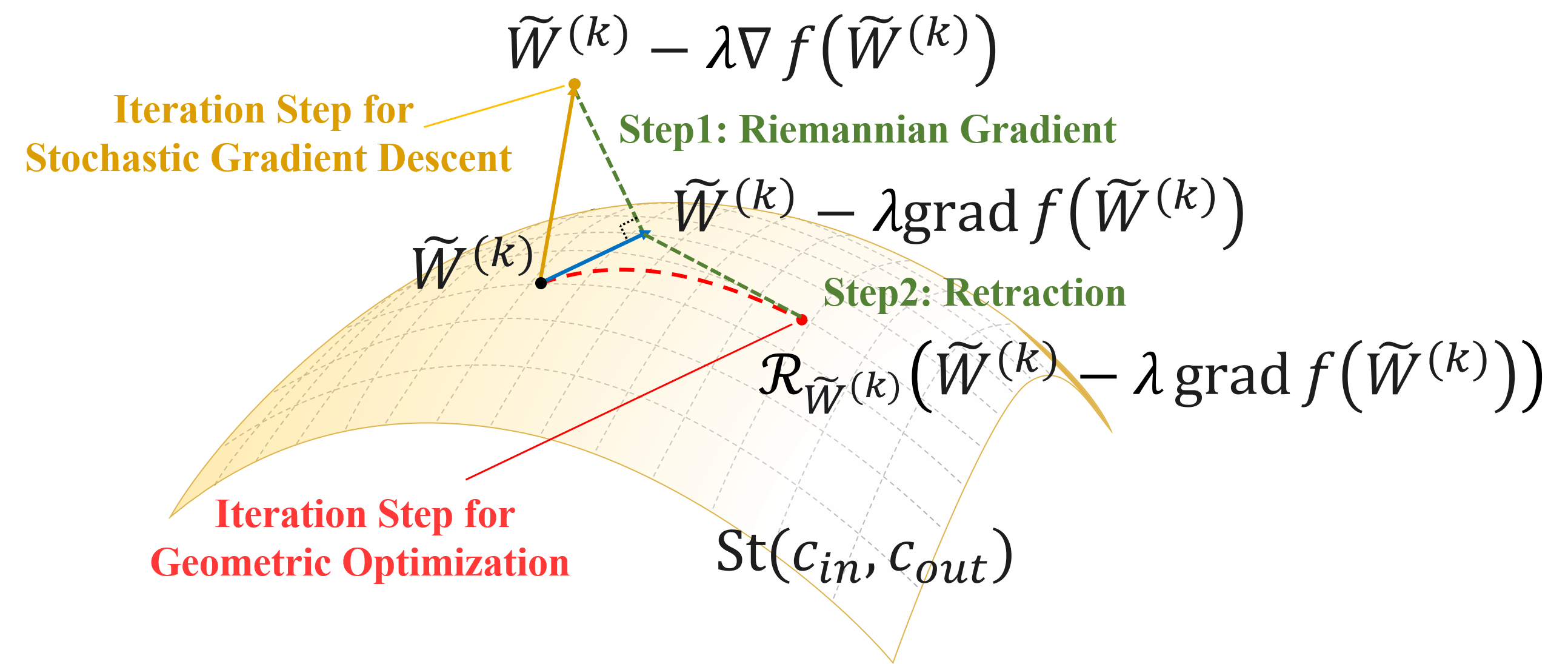}
\par\end{centering}
\vspace{-0.3cm}
\caption{\label{fig:visually-ill}Illustration of the geometric optimization process.}
\vspace{-0.2cm}
\end{figure}

Geometric optimization converts the constrained problem on Euclidean space to an unconstrained problem on the Stiefel manifold.
Stiefel manifolds express all sets of orthogonal matrices.
Let $\tilde{W}\in\textrm{St}(c_{in},c_{out})$ \textbf{$(c_{in}<c_{out})$} express the $c_{out}\times c_{in}$ Stiefel manifold, i.e.:
\begin{equation}
\ensuremath{\textrm{St}(c_{in},c_{out})=\left\{ \tilde{W}\in\mathbb{R}^{c_{out}\times c_{in}}:\ \tilde{W}^{T}\tilde{W}=I_{c_{in}}\right\} }.
\end{equation}
Let Euclidean manifold $\textrm{E}(c_{in},c_{out})$ be unconstrained Euclidean space of $c_{out}\times c_{in}$ matrices, $W\in\textrm{E}(c_{in},c_{out})$.
The constrained problem Eq. \eqref{eq:old-problem} can be converted to the unconstrained form:
\begin{equation}
\resizebox{0.42\textwidth}{!}{
\ensuremath{\begin{array}{c}
f:\textrm{St}(c_{in},c_{out})\times\textrm{E}(c_{in},c_{out})\rightarrow\mathbb{R},\\
f(W,\tilde{W}):=\left\Vert (GT-In)-\left(\tilde{W}\left(W^{T}(GT-In)\right)\right)\right\Vert ^{2}.
\end{array}}\label{eq:15}}
\end{equation}
Then, this unconstrained problem can be optimized by commonly used optimizers, such as SGD and Adam, but need to add two key steps, 
1) calculating the Riemannian gradient and 2) retraction transformation.
This process is visually illustrated in Fig. \ref{fig:visually-ill}.

The Riemannian gradient $\textrm{grad}f(X)$ can be calculated by:
\begin{equation}
\resizebox{0.42\textwidth}{!}{$
\textrm{grad}f(X)=\nabla f(X)-\frac{1}{2}XX{}^{T}\nabla f(X)-\frac{1}{2}X\nabla f(X)^{T}X.
$}
\label{eq:grad}
\end{equation}
where $\nabla f(\cdot)$ is the Euclidean gradient. The previous work \cite{ODCR} proved that the Riemannian gradient is the projection of the Euclidean gradient onto the tangent space, which means the steepest-ascent direction onto the Stiefel manifold.
Then, random mini-batches of training data is used to update the parameters along the negative direction of the gradient until convergence.
The updating process of $W$ takes the following form:
\begin{equation}
W^{(k+1)}=W^{(k)}-\lambda\nabla f(W^{(k)}),
\label{eq:update-normal}
\end{equation}
where $\lambda$ is the step size.
The updating process of $\tilde{W}$ takes the following form:
\begin{equation}
\tilde{W}^{(k+1)}=\mathcal{R}_{\tilde{W}^{(k)}}(\tilde{W}^{(k)}-\lambda\textrm{grad}f(\tilde{W}^{(k)})),
\label{eq:update-constraint}
\end{equation}
where $\mathcal{R}_{\tilde{W}^{(k)}}(\cdot)$ is the second key step: retraction transformation:
\begin{equation}
\mathcal{R}_{\tilde{W}^{(k)}}(\Xi)=(\tilde{W}^{(k)}+\Xi)\left(I_{c_{in}}+\Xi^{T}\Xi\right)^{-\frac{1}{2}}.
\label{eq:retraction}
\end{equation}
The previous work  \cite{SCA, ODCR} proved that after the retraction transformation, the iteration point satisfies the orthogonal constraint. \\

\noindent \textbf{Convergence Analysis.}
Although the Riemannian gradient has been proven to be the projection of the Euclidean gradient, it is necessary to theoretically prove its convergence considering the interaction between the adjusted search direction and the retraction function.
For convenience, we consider cost function $f$ and retraction transformation $\mathcal{R}$ as a whole and analyze the convergence of their composite function $g=f\circ\mathcal{R}(\cdot)$.

\noindent \textbf{Theorem 1}. Let $\tilde{W}$ update step by Eq. \eqref{eq:update-constraint}; the step size $\lambda$ is set to $\frac{1}{L}$, where $L$ is the constant from the $L$-smooth condition.
The function value decreases at each gradient descent step:
\begin{equation}
g(\tilde{W}^{(k+1)}) \leq g(\tilde{W}^{(k)}) - \frac{1}{2L} \|\textrm{grad}f(\tilde{W}^{(k)})\|^2.
\end{equation}\\
Before presenting the proof, we first introduce several lemmas.
The proofs of these lemmas are provided in Appendix B, please
refer to the supplementary material.\\

\noindent \textbf{Lemma 1}. Given the cost function $f(\cdot)$ defined in Eq. \eqref{eq:15} and the retraction transformation $\mathcal{R}_{\tilde{W}}(\cdot)$ defined in Eq. \eqref{eq:retraction}, the grads of $f(\cdot)$ equipped with or without $\mathcal{R}_{\tilde{W}}(\cdot)$ are the same.\\

\noindent \textbf{Lemma 2}. Given the cost function defined in Eq. \eqref{eq:15}, considering  $\tilde{W}$ as the variable, $f(\tilde{W})$ is $L$-smooth.\\

\noindent \textbf{Lemma 3}. Considering  $\tilde{W}$ as the variable, the composite function $g = f \circ \mathcal{R}(\tilde{W})$ is $L$-smooth.\\

\noindent \textit{Proof of Theorem 1. }
With the composite function $g=f\circ\mathcal{R}(\cdot)$ and $\lambda=\frac{1}{L}$, the update rule Eq. \eqref{eq:update-constraint} can be reformed as follows:
\begin{equation}
\tilde{W}^{(k+1)}=\tilde{W}^{(k)}-\frac{1}{L}\textrm{grad}f(\tilde{W}^{(k)}).
\label{eq:new_iter}
\end{equation}
Lemma 3 proves that the composite function $g=f\circ\mathcal{R}(\cdot)$ is $L$-smooth.
Since $g(\cdot)$ is $L$-smooth, we have:
\begin{equation}
g(Y) \leq g(X) + \langle \nabla g(X), Y - X \rangle + \frac{L}{2} \|Y - X\|^2.
\end{equation}
Let $X=\tilde{W}^{(k)}$ and $Y=\tilde{W}^{(k+1)}$, we get:
\begin{equation}
\begin{split}
g(\tilde{W}^{(k+1)}) \leq & g(\tilde{W}^{(k)}) + \\
& \langle \nabla g(\tilde{W}^{(k)}), \tilde{W}^{(k+1)} - \tilde{W}^{(k)} \rangle + \\
& \frac{L}{2} \|\tilde{W}^{(k+1)} - \tilde{W}^{(k)}\|^2.
\end{split}
\label{eq:lsmooth_property}
\end{equation}
From the update step Eq. \eqref{eq:new_iter}, we substitute into Eq. \eqref{eq:lsmooth_property}:
\begin{equation}
\frac{L}{2} \|\tilde{W}^{(k+1)} - \tilde{W}^{(k)}\|^2= \frac{1}{2L} \left\| \textrm{grad}f(\tilde{W}^{(k)}) \right\|^2.
\label{eq:last_item}
\end{equation}
Lemma 1 gives us the idea that the differential of the retraction transformation is identity mapping, i.e.:
\begin{equation}
\nabla g=\nabla f\circ \mathcal{R}=\nabla f.
\label{eq:identity_mapping}
\end{equation}
We substitute Eq. \eqref{eq:identity_mapping} into Eq. \eqref{eq:lsmooth_property}:
\begin{equation}
\resizebox{0.42\textwidth}{!}{$
\langle \nabla g(\tilde{W}^{(k)}), \tilde{W}^{(k+1)} - \tilde{W}^{(k)} \rangle = \langle \nabla f(\tilde{W}^{(k)}), \tilde{W}^{(k+1)} - \tilde{W}^{(k)} \rangle.
$}
\end{equation}
Please note that due to the retraction transformation applied to $g(\cdot)$, $ \tilde{W}^{(k+1)}$ is on the tangent space at $\tilde{W}^{(k)}$.
According to the previous work \cite{ODCR}, $\textrm{grad}f$ is the projection of $\nabla f$ into the tangent space. Therefore, we have:
\begin{equation}
\resizebox{0.42\textwidth}{!}{$
\langle \nabla g(\tilde{W}^{(k)}), \tilde{W}^{(k+1)} - \tilde{W}^{(k)} \rangle = -\frac{1}{L} \left\| \textrm{grad}f(\tilde{W}^{(k)}) \right\|^2.
$}
\label{eq:inner_product}
\end{equation}
Substitute Eq. \eqref{eq:last_item} and Eq. \eqref{eq:inner_product} into Eq. \eqref{eq:lsmooth_property}, we can simplify the expression:
\begin{equation}
g(\tilde{W}^{(k+1)}) \leq g(\tilde{W}^{(k)}) - \frac{1}{2L} \|\textrm{grad}f(\tilde{W}^{(k )})\|^2.
\end{equation}
We have proved that for the composite function $g=f\circ\mathcal{R}$, the function value decreases by at least $\frac{1}{2L} \|\textrm{grad}f(\tilde{W}^{(k )})\|^2$ at each gradient descent step.
After the iteration,  $g$ will converge to a local optimum.
$\hfill\blacksquare$

\subsection{Proof-of-Concept: Exposure Correction Model}

\begin{figure}[!t]
\hspace*{-0.05\columnwidth}
\includegraphics[width=1.05\columnwidth]{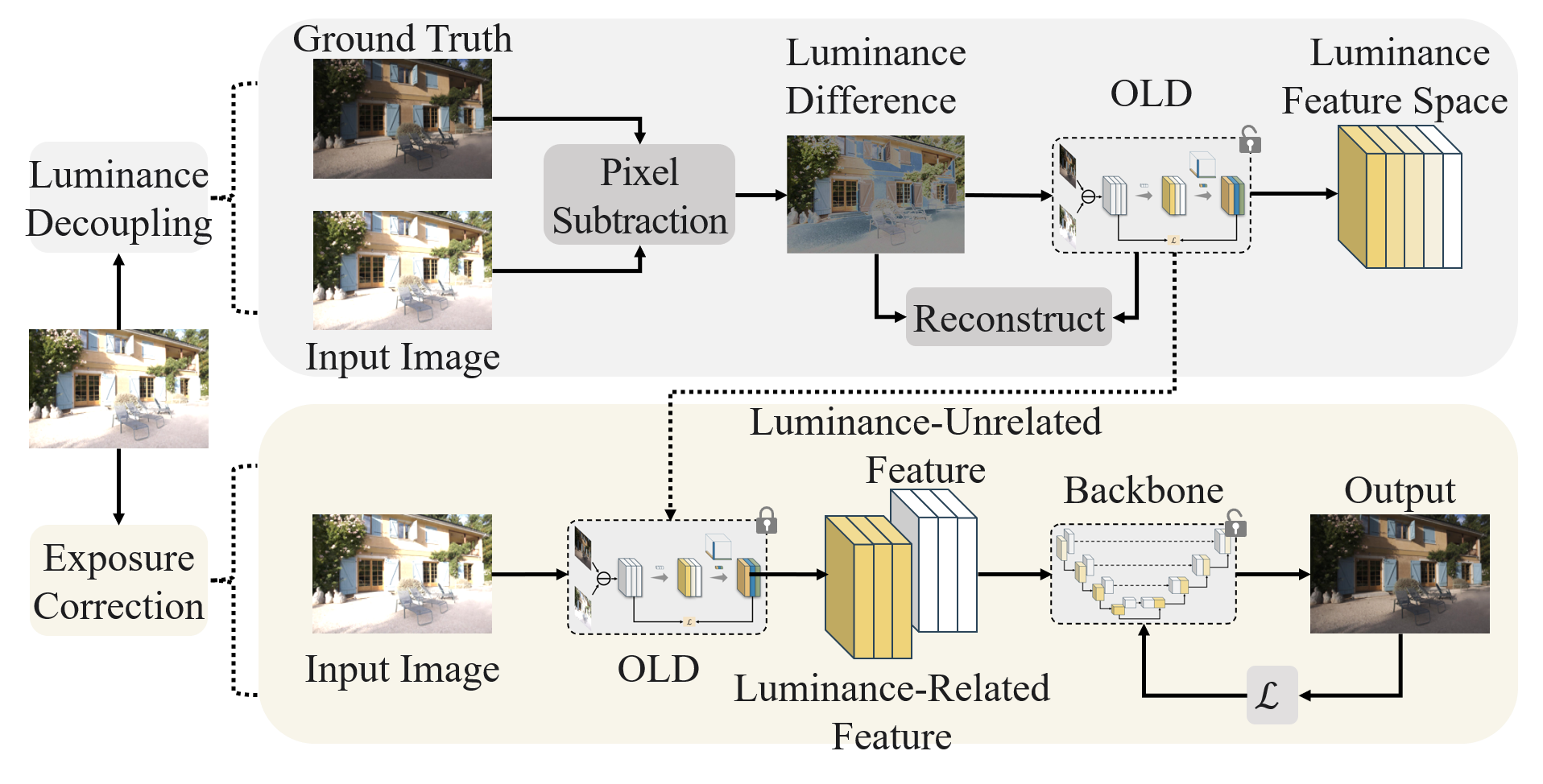}
\caption{The overall pipeline of the proposed LCA.
Images are processed by luminance decoupling and exposure correction.}
\label{fig:The-overall-str}
\end{figure}

For exposure correction we introduce a proof-of-concept model, called Luminance Component Analysis (LCA), which is illustrated in Fig. \ref{fig:The-overall-str}.
LCA uses a supervised setting, where image pairs $(In, GT)$ are provided for training.
These pairs consist of an input image with poor exposure ($In$) and a ground truth image with the correct exposure ($GT$).
LCA recursively decouples the luminance into the luminance-related components through the previously introduced OLD module.
Such that only the luminance-related components are adjusted to achieve precise exposure correction while preserving the integrity of color and texture in the luminance-unrelated components.
It is worth noting that the steps of luminance decoupling and exposure correcting are not separated but are integrated within the backbone, enabling cooperative searching in an end-to-end architecture.
The specific details of the backbone network, the integration of luminance decoupling and exposure correction, and the cooperatively searching methodology are provided in Appendix A, please refer to the supplementary material.

For the training loss, LCA attempts to model the conditional distribution of real images based on the input images by minimizing the adversarial loss:
\begin{equation}
\min_{Gen}\max_{Dis}\mathcal{L}_{GAN}(In,GT),
\end{equation}
where the adversarial loss $\mathcal{L}_{GAN}(In,GT)$ is given by:
\begin{equation}
\resizebox{0.42\textwidth}{!}{$
\mathcal{L}_{GAN}=E[\log D(In,GT)]+E[\log(1-D(In,G(In)))].
$}
\end{equation}
Additionally, to improve the similarity between the generated images or the ground truth through the well-trained neural networks, 
the perceptual loss, which uses VGG-19 to compare the L1 distance of the output image and ground truth in the feature space \cite{baseline}, 
replaces the pixel loss, which directly calculates the pixel difference between two images, in LCA:
\begin{equation}
\mathcal{L}_{per}=\sum_{i=1}^{5}\left\Vert VGG^{i}(GT)-VGG^{i}(G(In))\right\Vert,
\end{equation}
where $VGG^{i}(\cdot)$ means the feature outputted by the $i$-th
layer group of VGG-19. 

\begin{table*}[!t]
\begin{centering}
{\footnotesize{}\caption{The results of the exposure dataset.
The best results are highlighted with \textcolor{red}{red}, and the second-best results with \textcolor{blue}{blue}.
``-'' indicates that they did not provide this metric.
well- and over-exposure image results can be found in Appendix C. 
}
\label{tab:The-results-of}
}{\footnotesize\par}
\par\end{centering}
\centering{}\resizebox{1.00\textwidth}{!}{
\begin{tabular}{@{}l@{}lccccccccccccc}
\toprule 
\multirow{2}{*}{Group} & \multirow{2}{*}{\;Method} & \multicolumn{2}{c}{Expert A} & \multicolumn{2}{c}{Expert B} & \multicolumn{2}{c}{Expert C} & \multicolumn{2}{c}{Expert D} & \multicolumn{2}{c}{Expert E} & \multicolumn{2}{c}{Avg} & \multirow{2}{*}{PI}
\tabularnewline
\cmidrule{3-14} 
    &  & PSNR & SSIM & PSNR & SSIM & PSNR & SSIM & PSNR & SSIM & PSNR & SSIM & PSNR & SSIM
\tabularnewline

\midrule
\multirow{7}{*}{All} & DPE \cite{DPE} & 16.93 & 0.68 & 17.70 & 0.67 & 17.74 & 0.70 & 17.57 & 0.67 & 17.60 & 0.67 & 17.51 & 0.68 & 2.62
\tabularnewline
    & Zero-DCE \cite{Zero-DCE} & 11.64 & 0.54 & 12.56 & 0.54 & 12.06 & 0.54 & 12.96 & 0.55 & 13.77 & 0.58 & 12.60 & 0.55 & 2.87
\tabularnewline
    & Afifi et al. \cite{laplacian} & 19.11 & 0.74 & 19.96 & 0.72 & 20.08 & 0.76 & 18.87 & 0.71 & 18.86 & 0.72 & 19.38 & 0.73 & 2.25
\tabularnewline
    & PEC \cite{PEC} & 12.76 & 0.63 & 13.28 & 0.68 & 12.80 & 0.65 & 13.10 & 0.67 & 12.99 & 0.67 & 12.99 & 0.64 & -
\tabularnewline
    & ExReg \cite{multi-dimensional-regressor} & 20.20 & 0.76 & 21.77 & 0.80 & 22.05 & 0.79 & 19.93 & 0.77 & 19.99 & 0.77 & 20.79 & 0.78 & -
\tabularnewline
    & Eyiokur et al. \cite{baseline} & \textcolor{blue}{20.44} & \textcolor{blue}{0.86} & \textcolor{blue}{21.773} & \textcolor{blue}{0.89} & \textcolor{blue}{22.33} & \textcolor{red}{0.90} & \textcolor{blue}{19.98} & \textcolor{blue}{0.87} & \textcolor{blue}{19.84} & \textcolor{red}{0.88} & \textcolor{blue}{20.87} & \textcolor{red}{0.88} & \textcolor{blue}{2.24}
\tabularnewline
    & LCA (Ours) & \textcolor{red}{20.83} & \textcolor{red}{0.86} & \textcolor{red}{22.71} & \textcolor{red}{0.90} & \textcolor{red}{22.41} & \textcolor{blue}{0.88} & \textcolor{red}{20.55} & \textcolor{red}{0.87} & \textcolor{red}{20.17} & \textcolor{blue}{0.87} & \textcolor{red}{21.33} & \textcolor{red}{0.88} & \textcolor{red}{1.94}
\tabularnewline
\bottomrule
\end{tabular}}
\end{table*}

\begin{table*}[!t]
\caption{Ablation study of LCA on the exposure correction dataset.
\textquotedblleft\textsurd \textquotedblright{} indicates the technique is adopted, \textquotedblleft\texttimes \textquotedblright{} indicates not, and \textquotedblleft -\textquotedblright{} means not applicable.
}
\label{tab:Ablation-study-of-ld}
\centering{}\resizebox{0.98\textwidth}{!}{
\begin{tabular}{ccccccccc}
\toprule
\multirow{2}{*}{Network} & Depthwise Separable & Luminance Feature & Non-Luminance Feature & Computational & Number of & \multirow{2}{*}{FPS} & \multirow{2}{*}{PSNR} & \multirow{2}{*}{SSIM}
\tabularnewline
 & Convolution & Decoupling & Preserving & Complexity & Parameters &  &  & 
\tabularnewline
\midrule
Baseline & \texttimes{} & - & \texttimes{} & 67.08GMac & 213.86M & 23.07 & 20.66 & 0.85
\tabularnewline
\midrule
v1 & \texttimes{} & - & \textsurd{} & 17.76GMac & 73.99M & 28.53 & 20.41 & 0.86
\tabularnewline
v2 & \textsurd{} & \texttimes{} & \texttimes{} & 65.25GMac & 210.73M & 23.16 & 19.79 & 0.85
\tabularnewline
v3 & \textsurd{} & \texttimes{} & \textsurd{} & 17.31GMac & 73.21M & 28.80 & 20.25 & 0.85
\tabularnewline
v4 & \textsurd{} & \textsurd{} & \texttimes{} & 65.25GMac & 210.73M & 22.98 & 19.89 & 0.84
\tabularnewline
LCA & \textsurd{} & \textsurd{} & \textsurd{} & 17.31GMac & 73.21M & 28.72 & 20.83 & 0.86
\tabularnewline
\bottomrule
\end{tabular}}
\end{table*}

\section{Experiments}
In this section, we will introduce experimental settings, including datasets and evaluation metrics, and compare LCA with other methods.

\subsection{Comparison with Other Methods}
We train and test LCA on the PyTorch platform with an i7 10700K CPU and TITAN RTX GPU.
We use Adam to train LCA. The learning rate is 1e-4, and the number of epochs is 300. The scheduler is MultiStepLR, and the learning rate is reduced by half every 60 steps.
The exposure correction dataset proposed by Afifi et al. \cite{laplacian} is used to train LCA, which includes under-exposed, over-exposed, and well-exposed images.
The training set consists of 17,675 images, the validation set has 750 images, and the test set includes 5,905 images with five exposure values (+1.5, +1, 0, -1, -1.5 EVs).
To evaluate exposure correction performance with other methods, we utilize two commonly used metrics as described in the literature: peak signal-to-noise ratio (PSNR) and structural similarity index measure (SSIM).
PSNR measures the image quality of generated images.
SSIM calculates the structural similarity.

Tab. \ref{tab:The-results-of} presents the results of the exposure correction dataset. 
LCA obtains the highest PSNR in all cases, the highest SSIM in 4 cases, and the best PSNR and SSIM on average.
The complete table following Afifi et al. \cite{laplacian}, divided the table into three groups, is presented in Appendix C.
The second-best method (proposed by Eyiokur et al. \cite{baseline}) obtains the highest PSNR in 4 cases and the highest SSIM in 11 cases.
This is because the proposed method, LCA, clearly separates the luminance-related and luminance-unrelated features, minimizing the effect of changes on luminance-unrelated information.

\begin{figure*}[!t]
\begin{centering}
\subfloat[Change the luminance-related components.]{
\includegraphics[width=1\textwidth]{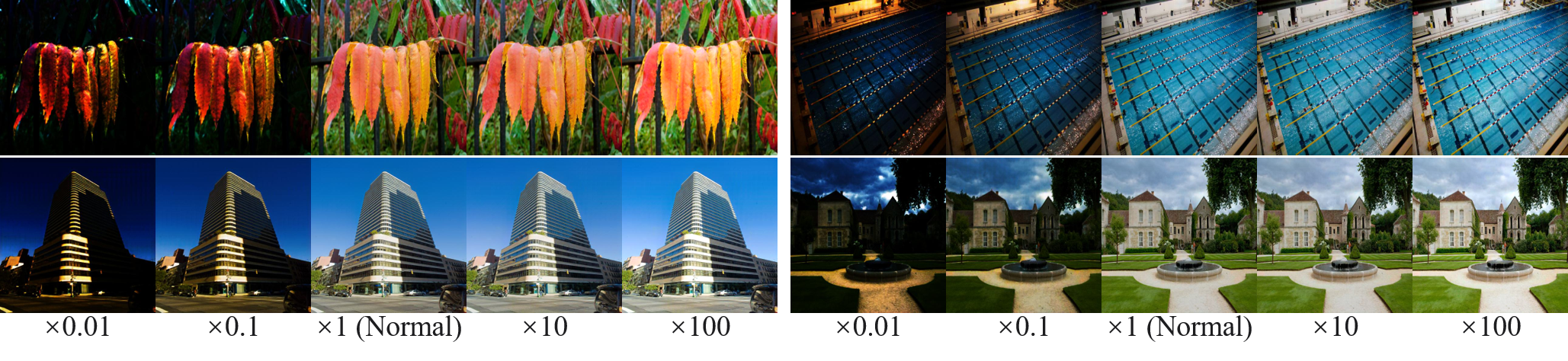}
}\par\end{centering}
\begin{centering}
\subfloat[Change the luminance-unrelated components.]{
\includegraphics[width=1\textwidth]{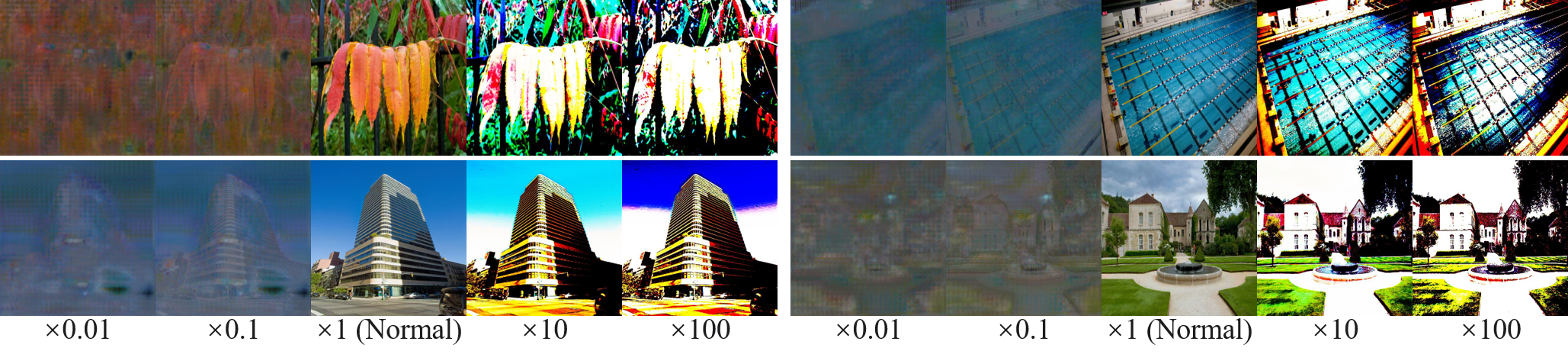}
}\par\end{centering}
\caption{Visualization results after changing the luminance-related and luminance-unrelated components.}
\label{fig:Visualization-results-after-ch}
\end{figure*}

\subsection{Ablation Study}
In order to validate the effectiveness of our major contributions, we implement three variations, including: 1) pruning the depthwise separable convolution, which is the prerequisite for OLD; 2) decoupling the luminance feature; 3) preserving the luminance-unrelated feature, as seen in Tab. \ref{tab:Ablation-study-of-ld}.
Among them, the luminance feature decoupling cannot be implemented without the depthwise separable convolution, so we exclude these methods. 

Compared baseline with v2, due to using the depthwise separable convolution, the number of parameters is largely reduced, and the performance is also decreased with 0.78 in PSNR.
But with the luminance feature decoupling and the luminance-unrelated feature preserving, LCA is the best; this demonstrates the effectiveness of decoupling the luminance feature in exposure correction.
Compared v4 with LCA, even though v4 also decouples the luminance feature, due to v4 not preserving the luminance-unrelated feature, its PSNR is 0.94 smaller than that of LCA.
In v1 and v3, the luminance-unrelated features cannot be kept because luminance-related and luminance-unrelated features are mixed up without luminance feature decoupling.

\subsection{Qualitative Performance}

\begin{figure}[!t]
\subfloat[SSIM.]{
\includegraphics[width=0.50\columnwidth]{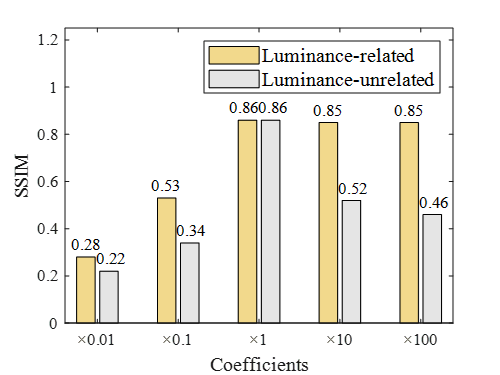}
}
\subfloat[Histogram matching score.]{
\includegraphics[width=0.50\columnwidth]{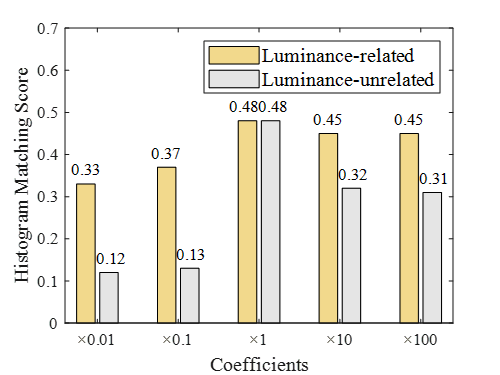}
}
\caption{Comparison of changing luminance-related or luminance-unrelated components. 
SSIM reflects the structural similarity, and the histogram matching score represents color distortion.
}
\label{fig:comparison-of-chan}
\end{figure}

To visually show whether the luminance-related and luminance-unrelated components decoupled by LCA indeed play their role (say the luminance-related components should be adjusted to correct the exposure value, while the luminance-unrelated components should be kept unchanged), we multiply different coefficients (0.01, 0.1, 10, and 100) with the luminance-related or luminance-unrelated components during the inference phase.
Fig. \ref{fig:Visualization-results-after-ch} shows the visualization results.
Among them, the results from Fig. \ref{fig:Visualization-results-after-ch} (a) are changing the luminance-related components, and Fig. \ref{fig:Visualization-results-after-ch} (b) shows results after altering the luminance-unrelated components.
The result shows that changing the luminance-related components can adjust the exposure value.
In contrast, any changes in the luminance-unrelated components would cause distortions in color and loss of detail.
The numerical comparison is presented in Fig. \ref{fig:comparison-of-chan}.
Although changing the well-calculated weights (whether luminance-related or luminance-unrelated components) reduces the SSIM and the histogram matching score,  changing the luminance-unrelated components can cause greater distortions in color and loss of structure than changing the luminance-related components.
Other experiments are presented in Appendix C.

\section{Conclusion}
This paper proposes a new method for exposure correction called luminance component analysis (LCA).
LCA combines orthogonal decomposition and deep learning, applying orthogonal decomposition to decouple the luminance-related and luminance-unrelated components.
LCA utilises a U-Net-like backbone and orthogonal luminance decouplers (OLD) to recursively decouple the luminance components.
The luminance-unrelated components remain unaltered, while only the luminance-related components are adjusted.
To optimize the problem with the orthogonal constraint, the geometric optimization algorithm converts the constrained problem in Euclidean space into the unconstrained problem in the orthogonal Stiefel manifold, which allows the existing mature optimization methods in deep learning to be used in LCA.
Crucially, this paper proves the convergence of geometric optimization, which laid the foundation for its extension to other fields.

Extensive experimentation demonstrated that LCA can decouple the luminance feature.
Additionally, LCA has achieved the highest PSNR score of 21.33 and SSIM score of 0.88 in the exposure correction dataset.
Future work will focus on a more efficient adjustment method for the luminance components with an effective network.


\bibliography{aaai25}

\begin{thebibliography}{43}
\providecommand{\natexlab}[1]{#1}

\bibitem[{Afifi et~al.(2021)Afifi, Derpanis, Ommer, and Brown}]{laplacian}
Afifi, M.; Derpanis, K.~G.; Ommer, B.; and Brown, M.~S. 2021.
\newblock Learning Multi-Scale Photo Exposure Correction.
\newblock In \emph{2021 IEEE/CVF Conference on Computer Vision and Pattern Recognition (CVPR)}, 9153--9163.

\bibitem[{Billmeyer~Jr.(1983)}]{color_space}
Billmeyer~Jr., F.~W. 1983.
\newblock Color Science: Concepts and Methods, Quantitative Data and Formulae, 2nd ed., by Gunter Wyszecki and W. S. Stiles, John Wiley and Sons, New York, 1982, 950 pp. Price: \$75.00.
\newblock \emph{Color Research \& Application}, 8(4): 262--263.

\bibitem[{Bull and Zhang(2021)}]{luminance_defination}
Bull, D.~R.; and Zhang, F. 2021.
\newblock Chapter 2 - The human visual system.
\newblock In Bull, D.~R.; and Zhang, F., eds., \emph{Intelligent Image and Video Compression (Second Edition)}, 17--58. Oxford: Academic Press, second edition edition.
\newblock ISBN 978-0-12-820353-8.

\bibitem[{Chen et~al.(2018)Chen, Wang, Kao, and Chuang}]{DPE}
Chen, Y.-S.; Wang, Y.-C.; Kao, M.-H.; and Chuang, Y.-Y. 2018.
\newblock Deep Photo Enhancer: Unpaired Learning for Image Enhancement from Photographs with GANs.
\newblock In \emph{2018 IEEE/CVF Conference on Computer Vision and Pattern Recognition}, 6306--6314.

\bibitem[{Chen et~al.(2016)Chen, Wen, Yao, and Jiang}]{image_processing_techniques2}
Chen, Z.; Wen, Y.; Yao, D.; and Jiang, B. 2016.
\newblock {A novel method to real-time offset correction for frame transfer CCD}.
\newblock In Jiang, Y.; Kippelen, B.; and Yu, J., eds., \emph{8th International Symposium on Advanced Optical Manufacturing and Testing Technologies: Optoelectronic Materials and Devices}, volume 9686, 96861L. International Society for Optics and Photonics, SPIE.

\bibitem[{Chiang et~al.(2022)Chiang, Hsueh, Hsiao, and Huang}]{multi-dimensional-regressor}
Chiang, T.-H.; Hsueh, H.-C.; Hsiao, C.-C.; and Huang, C.-C. 2022.
\newblock ExReg: Wide-range Photo Exposure Correction via a Multi-dimensional Regressor with Attention.
\newblock \emph{ArXiv}, abs/2212.14801.

\bibitem[{Cui et~al.(2022)Cui, Li, Gu, Su, Gao, Jiang, Qiao, and Harada}]{You_Only_Need_90K_P}
Cui, Z.; Li, K.; Gu, L.; Su, S.; Gao, P.; Jiang, Z.; Qiao, Y.; and Harada, T. 2022.
\newblock You Only Need 90K Parameters to Adapt Light: a Light Weight Transformer for Image Enhancement and Exposure Correction.
\newblock In \emph{33rd British Machine Vision Conference 2022, {BMVC} 2022, London, UK, November 21-24, 2022}. {BMVA} Press.

\bibitem[{Dacey(2000)}]{HVS3}
Dacey, D.~M. 2000.
\newblock Parallel Pathways for Spectral Coding in Primate Retina.
\newblock \emph{Annual Review of Neuroscience}, 23(Volume 23, 2000): 743--775.

\bibitem[{Eyiokur et~al.(2022)Eyiokur, Yaman, Ekenel, and Waibel}]{baseline}
Eyiokur, F.~I.; Yaman, D.; Ekenel, H.~K.; and Waibel, A. 2022.
\newblock Exposure Correction Model to Enhance Image Quality.
\newblock In \emph{2022 IEEE/CVF Conference on Computer Vision and Pattern Recognition Workshops (CVPRW)}, 675--685.

\bibitem[{Fei et~al.(2023)Fei, Wei, Liu, Li, and Chen}]{manifold_survey}
Fei, Y.; Wei, X.; Liu, Y.; Li, Z.; and Chen, M. 2023.
\newblock A Survey of Geometric Optimization for Deep Learning: From Euclidean Space to Riemannian Manifold.
\newblock \emph{ArXiv}, abs/2302.08210.

\bibitem[{Feng et~al.(2021)Feng, Li, Hua, and Zhang}]{detail_useless}
Feng, X.; Li, J.; Hua, Z.; and Zhang, F. 2021.
\newblock Low-light image enhancement based on multi-illumination estimation.
\newblock \emph{Applied Intelligence}, 51(7): 5111--5131.

\bibitem[{Finlayson, Mohammadzadeh~Darrodi, and Mackiewicz(2015)}]{color_correction_matrix}
Finlayson, G.~D.; Mohammadzadeh~Darrodi, M.; and Mackiewicz, M. 2015.
\newblock The alternating least squares technique for nonuniform intensity color correction.
\newblock \emph{Color Research \& Application}, 40(3): 232--242.

\bibitem[{Guo et~al.(2020)Guo, Li, Guo, Loy, Hou, Kwong, and Cong}]{Zero-DCE}
Guo, C.; Li, C.; Guo, J.; Loy, C.~C.; Hou, J.; Kwong, S.; and Cong, R. 2020.
\newblock Zero-Reference Deep Curve Estimation for Low-Light Image Enhancement.
\newblock In \emph{2020 IEEE/CVF Conference on Computer Vision and Pattern Recognition (CVPR)}, 1777--1786.

\bibitem[{Huang, Hung, and Chen(2022)}]{orthogonal_constraint1}
Huang, G.-Y.; Hung, C.-Y.; and Chen, B.-W. 2022.
\newblock Image feature selection based on orthogonal l2,0 norms.
\newblock \emph{Measurement}, 199: 111310.

\bibitem[{Huang et~al.(2022)Huang, Liu, Zhao, Yan, Zhang, Huang, Zhou, and Xiong}]{Fourier}
Huang, J.; Liu, Y.; Zhao, F.; Yan, K.; Zhang, J.; Huang, Y.; Zhou, M.; and Xiong, Z. 2022.
\newblock Deep Fourier-Based Exposure Correction Network with Spatial-Frequency Interaction.
\newblock In Avidan, S.; Brostow, G.; Ciss{\'e}, M.; Farinella, G.~M.; and Hassner, T., eds., \emph{Computer Vision -- ECCV 2022}, 163--180. Cham: Springer Nature Switzerland.
\newblock ISBN 978-3-031-19800-7.

\bibitem[{Jr. and Peterson(1992)}]{color_compression_luminance}
Jr., A. J.~A.; and Peterson, H.~A. 1992.
\newblock {Luminance-model-based DCT quantization for color image compression}.
\newblock In Rogowitz, B.~E., ed., \emph{Human Vision, Visual Processing, and Digital Display III}, volume 1666, 365 -- 374. International Society for Optics and Photonics, SPIE.

\bibitem[{Kang et~al.(2011)Kang, Jeon, Han, and Ko}]{YCbCr}
Kang, B.; Jeon, C.; Han, D.~K.; and Ko, H. 2011.
\newblock Adaptive height-modified histogram equalization and chroma correction in YCbCr color space for fast backlight image compensation.
\newblock \emph{Image and Vision Computing}, 29(8): 557--568.

\bibitem[{Kennedy(2010)}]{HVS2}
Kennedy, G.~J. 2010.
\newblock Visual Perception: A Clinical Orientation.
\newblock \emph{Ophthalmic and Physiological Optics}, 30(4): 408--408.

\bibitem[{Klein, Silverstein, and Carney(1992)}]{JPEG}
Klein, S.~A.; Silverstein, D.~A.; and Carney, T. 1992.
\newblock {Relevance of human vision to JPEG-DCT compression}.
\newblock In Rogowitz, B.~E., ed., \emph{Human Vision, Visual Processing, and Digital Display III}, volume 1666, 200 -- 215. International Society for Optics and Photonics, SPIE.

\bibitem[{Kumar, Jha, and Nishchal(2021)}]{HSV}
Kumar, A.; Jha, R.~K.; and Nishchal, N.~K. 2021.
\newblock An improved Gamma correction model for image dehazing in a multi-exposure fusion framework.
\newblock \emph{Journal of Visual Communication and Image Representation}, 78: 103122.

\bibitem[{Li, Li, and Todorovic(2020)}]{stiefel_manifold_SGD}
Li, J.; Li, F.; and Todorovic, S. 2020.
\newblock Efficient Riemannian Optimization on the Stiefel Manifold via the Cayley Transform.
\newblock In \emph{International Conference on Learning Representations}.

\bibitem[{Li et~al.(2023)Li, Li, Wang, Jiao, and Bian}]{DecomNet}
Li, L.; Li, D.; Wang, S.; Jiao, Q.; and Bian, L. 2023.
\newblock Tuning-free and self-supervised image enhancement against ill exposure.
\newblock \emph{Opt. Express}, 31(6): 10368--10385.

\bibitem[{Li et~al.(2018)Li, Liu, Yang, Sun, and Guo}]{Retinex3}
Li, M.; Liu, J.; Yang, W.; Sun, X.; and Guo, Z. 2018.
\newblock Structure-Revealing Low-Light Image Enhancement Via Robust Retinex Model.
\newblock \emph{IEEE Transactions on Image Processing}, 27(6): 2828--2841.

\bibitem[{Li, Fang, and Liu(2021)}]{orthogonal_constraint3}
Li, X.; Fang, M.; and Liu, J. 2021.
\newblock Low-rank embedded orthogonal subspace learning for zero-shot classification.
\newblock \emph{Journal of Visual Communication and Image Representation}, 74: 102981.

\bibitem[{Liu et~al.(2023)Liu, Wu, Luan, Jiang, Liu, and Fan}]{HoLoCo}
Liu, J.; Wu, G.; Luan, J.; Jiang, Z.; Liu, R.; and Fan, X. 2023.
\newblock HoLoCo: Holistic and local contrastive learning network for multi-exposure image fusion.
\newblock \emph{Information Fusion}, 95: 237--249.

\bibitem[{Liu et~al.(2021)Liu, Ma, Zhang, Fan, and Luo}]{RUAS}
Liu, R.; Ma, L.; Zhang, J.; Fan, X.; and Luo, Z. 2021.
\newblock Retinex-inspired Unrolling with Cooperative Prior Architecture Search for Low-light Image Enhancement.
\newblock In \emph{2021 IEEE/CVF Conference on Computer Vision and Pattern Recognition (CVPR)}, 10556--10565.

\bibitem[{Ma et~al.(2022{\natexlab{a}})Ma, Ma, Liu, Fan, and Luo}]{SCI}
Ma, L.; Ma, T.; Liu, R.; Fan, X.; and Luo, Z. 2022{\natexlab{a}}.
\newblock Toward Fast, Flexible, and Robust Low-Light Image Enhancement.
\newblock In \emph{2022 IEEE/CVF Conference on Computer Vision and Pattern Recognition (CVPR)}, 5627--5636.

\bibitem[{Ma et~al.(2022{\natexlab{b}})Ma, Ma, Xue, Fan, Luo, and Liu}]{PEC}
Ma, L.; Ma, T.; Xue, X.; Fan, X.; Luo, Z.; and Liu, R. 2022{\natexlab{b}}.
\newblock Practical Exposure Correction: Great Truths Are Always Simple.
\newblock \emph{ArXiv}, abs/2212.14245.

\bibitem[{Nathans et~al.(1986)Nathans, Piantanida, Eddy, Shows, and Hogness}]{HVS1}
Nathans, J.; Piantanida, T.~P.; Eddy, R.~L.; Shows, T.~B.; and Hogness, D.~S. 1986.
\newblock Molecular Genetics of Inherited Variation in Human Color Vision.
\newblock \emph{Science}, 232(4747): 203--210.

\bibitem[{Peng, Zhao, and Hu(2021)}]{SCA}
Peng, J.; Zhao, H.; and Hu, Z. 2021.
\newblock Second-order component analysis for fault detection.
\newblock \emph{Journal of Process Control}, 108: 25--39.

\bibitem[{Peng, Liao, and Chen(2022)}]{gamma_correction}
Peng, Y.-T.; Liao, H.-H.; and Chen, C.-F. 2022.
\newblock Two-Exposure Image Fusion Based on Optimized Adaptive Gamma Correction.
\newblock \emph{Sensors}, 22(1).

\bibitem[{Rahman et~al.(2020)Rahman, Yi-Fei, Aamir, Wali, and Guan}]{traditional_methods_not_work}
Rahman, Z.; Yi-Fei, P.; Aamir, M.; Wali, S.; and Guan, Y. 2020.
\newblock Efficient Image Enhancement Model for Correcting Uneven Illumination Images.
\newblock \emph{IEEE Access}, 8: 109038--109053.

\bibitem[{Ren et~al.(2020)Ren, Yang, Cheng, and Liu}]{Retinex1}
Ren, X.; Yang, W.; Cheng, W.-H.; and Liu, J. 2020.
\newblock LR3M: Robust Low-Light Enhancement via Low-Rank Regularized Retinex Model.
\newblock \emph{IEEE Transactions on Image Processing}, 29: 5862--5876.

\bibitem[{Torres et~al.(2015)Torres, Schutte, Bouma, and Menéndez}]{image_processing_techniques1}
Torres, J.; Schutte, K.; Bouma, H.; and Menéndez, J.-M. 2015.
\newblock Linear colour correction for multiple illumination changes and non-overlapping cameras.
\newblock \emph{IET Image Processing}, 9(4): 280--289.

\bibitem[{Wang et~al.(2019)Wang, Zhang, Fu, Shen, Zheng, and Jia}]{Deep_UPE}
Wang, R.; Zhang, Q.; Fu, C.-W.; Shen, X.; Zheng, W.-S.; and Jia, J. 2019.
\newblock Underexposed Photo Enhancement Using Deep Illumination Estimation.
\newblock In \emph{2019 IEEE/CVF Conference on Computer Vision and Pattern Recognition (CVPR)}, 6842--6850.

\bibitem[{Wang et~al.(2018)Wang, Liu, Zhu, Tao, Kautz, and Catanzaro}]{pix2pixHD}
Wang, T.-C.; Liu, M.-Y.; Zhu, J.-Y.; Tao, A.; Kautz, J.; and Catanzaro, B. 2018.
\newblock High-Resolution Image Synthesis and Semantic Manipulation with Conditional GANs.
\newblock In \emph{2018 IEEE/CVF Conference on Computer Vision and Pattern Recognition}, 8798--8807.

\bibitem[{Wang et~al.(2024)Wang, Zhao, Peng, Yao, and Zhao}]{ODCR}
Wang, Z.; Zhao, H.; Peng, J.; Yao, L.; and Zhao, K. 2024.
\newblock ODCR: Orthogonal Decoupling Contrastive Regularization for Unpaired Image Dehazing.
\newblock In \emph{Proceedings of the IEEE/CVF Conference on Computer Vision and Pattern Recognition (CVPR)}, 25479--25489.

\bibitem[{Wei et~al.(2018)Wei, Wang, Yang, and Liu}]{RetinexNet}
Wei, C.; Wang, W.; Yang, W.; and Liu, J. 2018.
\newblock Deep Retinex Decomposition for Low-Light Enhancement.
\newblock In \emph{British Machine Vision Conference}. British Machine Vision Association.

\bibitem[{Yadav et~al.(2021)Yadav, Ghosal, Lutz, and Smolic}]{discrete_cosine}
Yadav, O.; Ghosal, K.; Lutz, S.; and Smolic, A. 2021.
\newblock Frequency-domain loss function for deep exposure correction of dark images.
\newblock \emph{Signal, Image and Video Processing}, 15(8): 1829--1836.

\bibitem[{Yang et~al.(2023)Yang, Cheng, Zhao, Yan, Zhang, and Li}]{LA-Net}
Yang, K.-F.; Cheng, C.; Zhao, S.-X.; Yan, H.-M.; Zhang, X.-S.; and Li, Y.-J. 2023.
\newblock Learning to Adapt to Light.
\newblock \emph{International Journal of Computer Vision}, 131(4): 1022--1041.

\bibitem[{Ying et~al.(2017)Ying, Li, Ren, Wang, and Wang}]{histogram3}
Ying, Z.; Li, G.; Ren, Y.; Wang, R.; and Wang, W. 2017.
\newblock A New Image Contrast Enhancement Algorithm Using Exposure Fusion Framework.
\newblock In Felsberg, M.; Heyden, A.; and Kr{\"u}ger, N., eds., \emph{Computer Analysis of Images and Patterns}, 36--46. Cham: Springer International Publishing.
\newblock ISBN 978-3-319-64698-5.

\bibitem[{Zhang et~al.(2018)Zhang, Yuan, Xiao, Zhu, and Zheng}]{Retinex2}
Zhang, Q.; Yuan, G.; Xiao, C.; Zhu, L.; and Zheng, W.-S. 2018.
\newblock High-Quality Exposure Correction of Underexposed Photos.
\newblock In \emph{Proceedings of the 26th ACM International Conference on Multimedia}, MM '18, 582–590. New York, NY, USA: Association for Computing Machinery.
\newblock ISBN 9781450356657.

\bibitem[{Zhao et~al.(2020)Zhao, Lai, Leung, and Zhang}]{orthogonal_constraint2}
Zhao, H.; Lai, Z.; Leung, H.; and Zhang, X. 2020.
\newblock \emph{Sparse Feature Learning}, 103--133.
\newblock Cham: Springer International Publishing.
\newblock ISBN 978-3-030-40794-0.

\end{thebibliography}

\newpage
\section*{Appendix A. Architecture Details}
\begin{figure}[!b]
\includegraphics[width=1.0\columnwidth]{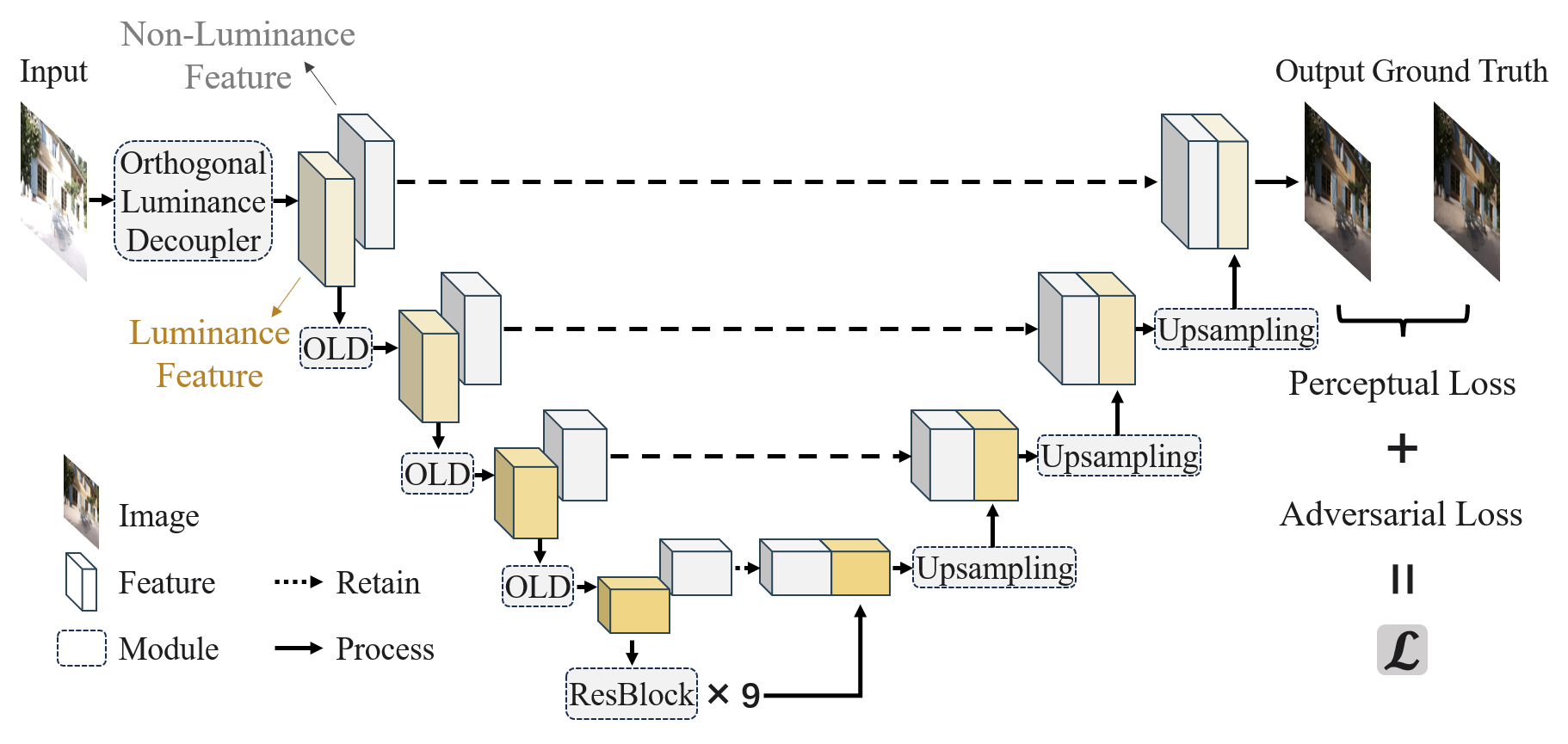}
\caption{\label{fig:Images-are-processed-b}The overall structure of LCA.}
\end{figure}

LCA consists of two modules: a generator and a discriminator.
The generator, based on a U-Net structure with nine added ResNet blocks for improved performance on high-resolution images \cite{pix2pixHD}, produces properly exposed images from poorly exposed ones.
The overall structure is illustrated in Fig. \ref{fig:Images-are-processed-b}.
The discriminator, a fully convolutional network, distinguishes real images from generated ones by channel-wise concatenating inputs with either generated or ground truth images.

The detailed network structure of LCA is shown in Tab. \ref{tab:Architectural-Details-of}.
Modules in the second column are used only during training, while those in merged cells are used in both the training and inference phases.
During training, LCA inputs both the original image (for calculating perceptual and adversarial losses) and the difference between the input and ground truth images (for training the OLD module). 
In the inference phase, LCA processes only the input image, as no ground truth is provided.
The optimizing steps of OLD and other parts of LCA are not separated but work cooperatively within an end-to-end architecture, as outlined in Alg. \ref{alg:algorithm}. 

\begin{table}[!t]
\caption{Architectural details of LCA.}
\label{tab:Architectural-Details-of}
\noindent\resizebox{0.50\textwidth}{!}{
\begin{tabular}{|cc|}
\hline
\multicolumn{2}{|c|}{(Generator)}                                                                           \\ \hline
\multicolumn{1}{|c|}{Input Image   (training and inference)}   & Input Image$-$Ground Truth   (training)    \\ \hline
\multicolumn{2}{|c|}{conv3$\rightarrow$64}                                                                  \\ \hline
\multicolumn{2}{|c|}{1\texttimes 1 conv64$\rightarrow$64,   BN, ReLU}                                       \\ \hline
\multicolumn{1}{|c|}{}                             & 1\texttimes 1 conv64$\rightarrow$64,   BN, sigmoid     \\ \hline
\multicolumn{2}{|c|}{conv32$\rightarrow$32 (groups=32)}                                                     \\ \hline
\multicolumn{2}{|c|}{1\texttimes 1 conv32$\rightarrow$128,   BN, ReLU}                                      \\ \hline
\multicolumn{1}{|c|}{}                             & 1\texttimes 1 conv128$\rightarrow$32,   BN, sigmoid    \\ \hline
\multicolumn{2}{|c|}{conv64$\rightarrow$64 (groups=64)}                                                     \\ \hline
\multicolumn{2}{|c|}{1\texttimes 1 conv64$\rightarrow$256,   BN, ReLU}                                      \\ \hline
\multicolumn{1}{|c|}{}                             & 1\texttimes 1 conv256$\rightarrow$64,   BN, sigmoid    \\ \hline
\multicolumn{2}{|c|}{conv128$\rightarrow$128 (groups=128)}                                                  \\ \hline
\multicolumn{2}{|c|}{1\texttimes 1 conv128$\rightarrow$512,   BN, ReLU}                                     \\ \hline
\multicolumn{1}{|c|}{}                             & 1\texttimes 1 conv512$\rightarrow$128,   BN, sigmoid   \\ \hline
\multicolumn{2}{|c|}{conv256$\rightarrow$256 (groups=256)}                                                  \\ \hline
\multicolumn{2}{|c|}{1\texttimes 1   conv256$\rightarrow$1024, BN, ReLU}                                    \\ \hline
\multicolumn{1}{|c|}{}                             & 1\texttimes 1 conv1024$\rightarrow$256,   BN, sigmoid  \\ \hline
\multicolumn{2}{|c|}{{[}conv512$\rightarrow$512,   BN, ReLU, conv512$\rightarrow$512, BN{]}\texttimes 9}    \\ \hline
\multicolumn{2}{|c|}{convTranspose1024$\rightarrow$256,   BN, ReLU}                                         \\ \hline
\multicolumn{2}{|c|}{convTranspose512$\rightarrow$128,   BN, ReLU}                                          \\ \hline
\multicolumn{2}{|c|}{convTranspose256$\rightarrow$64,   BN, ReLU}                                           \\ \hline
\multicolumn{2}{|c|}{convTranspose128$\rightarrow$32,   BN, ReLU}                                           \\ \hline
\multicolumn{2}{|c|}{conv64$\rightarrow$3, tanh}                                                            \\ \hline
\multicolumn{1}{|c|}{\multirow{7}{*}{}}            & (Discriminator)                                        \\ \cline{2-2} 
\multicolumn{1}{|c|}{}                             & conv6$\rightarrow$64, ReLU                             \\ \cline{2-2} 
\multicolumn{1}{|c|}{}                             & conv64$\rightarrow$128, BN, ReLU                       \\ \cline{2-2} 
\multicolumn{1}{|c|}{}                             & conv128$\rightarrow$256, BN, ReLU                      \\ \cline{2-2} 
\multicolumn{1}{|c|}{}                             & conv256$\rightarrow$512, BN, ReLU                      \\ \cline{2-2} 
\multicolumn{1}{|c|}{}                             & conv512$\rightarrow$512, BN, ReLU                      \\ \cline{2-2} 
\multicolumn{1}{|c|}{}                             & conv512$\rightarrow$1                                  \\ \hline
\end{tabular}}
\end{table}

\begin{algorithm}[!b]
\small
\caption{\label{alg:algorithm}Cooperative Architecture Search Strategy} 
\hspace*{0.02in} {\noindent \bf Input:}
The loss function $\mathcal{L}(\cdot)$ and the training dataset.\\
\hspace*{0.02in} {\noindent \bf Parameter:}
Parameters of OLD ($\tilde{W}^{(k)}$) and other parts ($W^{(k)}$).\\
\hspace*{0.02in} {\noindent \bf Output:}
The converged $W^{(k)}$ and $\tilde{W}^{(k)}$.

\begin{algorithmic}
\State Initialize $W^{(0)}$ and $\tilde{W}^{(0)}$.
\While {not converged}
\State // Update parameters for the main model.
\State Update $W^{(k)}$ through the back-propagation algorithm:
\begin{equation}
    W^{(k+1)}=W^{(k)}-\lambda\nabla f(W^{(k)}),
\end{equation}
 where $\lambda$ is the learning rate.
\State // Update parameters for OLD.
\State Update $\tilde{W}^{(k)}$ through the geometric optimization:
\begin{equation}
\tilde{W}^{(k+1)}=\mathcal{R}(\tilde{W}^{(k)}-\lambda\textrm{grad}f(\tilde{W}^{(k)})),
\end{equation}
where $\textrm{grad}(\cdot)$ refers to Riemannian gradient and $\mathcal{R}(\cdot)$ means retraction transformation.
\EndWhile
\State \textbf{return} $W^{(k)}$ and $\tilde{W}^{(k)}$.
\end{algorithmic}
\end{algorithm}

\clearpage
\onecolumn
\section*{Appendix B. Proofs of Lemmas.}

\noindent \textbf{Lemma 1}. Given the cost function $f(\cdot)$ and the retraction transformation $\mathcal{R}_{\tilde{W}}(\cdot)$ at the point of $\tilde{W}$ defined in Section 3 Methdology, the grads of $f(\cdot)$ equipped with or without $\mathcal{R}_{\tilde{W}}(\cdot)$ are the same.\\

\noindent \textit{Proof.} Consider $\mathcal{R}_{\tilde{W}}(t\xi)$
\begin{equation}
\mathcal{R}_{\tilde{W}}(t\xi) = ({\tilde{W}} + t\xi)(I_{c_{in}} + t^2 \xi^T \xi)^{-\frac{1}{2}},
\end{equation}
the derivative with respect to $t$ at $t=0$ is:
\begin{equation}
\frac{d}{dt} \mathcal{R}_{\tilde{W}}(t\xi) \Big|_{t=0} = \frac{d}{dt} \left[ ({\tilde{W}} + t\xi)(I_{c_{in}} + t^2 \xi^T \xi)^{-\frac{1}{2}} \right] \Big|_{t=0}.
\end{equation}
Let $A(t)$ denote $(I_{c_{in}} + t^2 \xi^T \xi)^{-\frac{1}{2}}$, we have:
\begin{equation}
\frac{d}{dt} A(t) \Big|_{t=0} = -\frac{1}{2} (I_{c_{in}} + t^2 \xi^T \xi)^{-\frac{3}{2}} 2t \xi^T \xi = 0.
\label{eq:23}
\end{equation}
Now, considering the product rule for the derivative:
\begin{equation}
\frac{d}{dt} \mathcal{R}_{\tilde{W}}(t\xi) \Big|_{t=0} = \frac{d}{dt} ({\tilde{W}} + t\xi) \Big|_{t=0} A(0) + ({\tilde{W}} + t\xi) \frac{d}{dt} A(t) \Big|_{t=0}.
\end{equation}
The first term is $\xi$ since $\frac{d}{dt} ({\tilde{W}} + t\xi) = \xi$ and $A(0) = I_{c_{in}}$.
The second term is zero, as shown from Eq. \eqref{eq:23}.
Therefore:
\begin{equation}
\frac{d}{dt} \mathcal{R}_{\tilde{W}}(t\xi) \Big|_{t=0} = \xi.
\end{equation}
That means the differential of the retraction transformation is identity mapping.
So the grad of $f(\cdot)$ equipped with $\mathcal{R}_{\tilde{W}}(\cdot)$ follows that:
\begin{equation}
\textrm{grad}f\circ \mathcal{R}_{\tilde{W}}=\textrm{grad}f,
\label{eq:21}
\end{equation}
which means the grads of $f(\cdot)$ equipped with or without $\mathcal{R}_{\tilde{W}}(\cdot)$ are the same.
$\hfill\blacksquare$\\

\noindent \textbf{Lemma 2}. $f(\tilde{W}) := \left\Vert (GT-In) - \left( \tilde{W} (W^{T}(GT-In)) \right) \right\Vert^2$ is $L$-smooth.\\

\noindent \textit{Proof.}
Let $A=GT-In$ and $B=W^T A$. The function simplifies to:
\begin{equation}
f(\tilde{W}) = \|A - \tilde{W} B\|^2.
\end{equation}
The gradient with respect to $\tilde{W}$ can be calculated by:
\begin{equation}
\nabla f(\tilde{W}) = -2 (A - \tilde{W} B) (B^T).
\end{equation}
Given two different points $\tilde{W}_1$ and $\tilde{W}_2$, their gradient is:
\begin{equation}
\begin{split}
\nabla f(\tilde{W}_1) = -2 (A - \tilde{W}_1 B) B^T,\\
\nabla f(\tilde{W}_2) = -2 (A - \tilde{W}_2 B) B^T.
\end{split}
\end{equation}
Their difference is:
\begin{equation}
\begin{split}
\nabla f(\tilde{W}_1) - \nabla f(\tilde{W}_2) 
&= -2 \left((A - \tilde{W}_1 B) B^T - (A - \tilde{W}_2 B) B^T\right) \\
&= -2 \left((\tilde{W}_2 B - \tilde{W}_1 B) B^T\right) \\
&= -2 (\tilde{W}_2 - \tilde{W}_1) B B^T.
\end{split}
\end{equation}
Therefore,
\begin{equation}
\|\nabla f(\tilde{W}_1) - \nabla f(\tilde{W}_2)\| = 2 \|(\tilde{W}_2 - \tilde{W}_1) B B^T\|.
\end{equation}
Using the properties of matrix norms, we further obtain:
\begin{equation}
\|\nabla f(\tilde{W}_1) - \nabla f(\tilde{W}_2)\| \leq 2 \|B B^T\| \|\tilde{W}_2 - \tilde{W}_1\|.
\end{equation}
Since $\|B B^T\|$ is a constant (because $B = W^T A$, where $A$ is a constant matrix and $W$ is independent of $\tilde{W}$), we can let $L = 2 \|B B^T\|$.
Thus, we have shown that there exists a constant $L = 2 \|B B^T\|$ such that for all orthogonal matrices $\tilde{W}_1$ and $\tilde{W}_2$:
\begin{equation}
\|\nabla f(\tilde{W}_1) - \nabla f(\tilde{W}_2)\| \leq L \|\tilde{W}_1 - \tilde{W}_2\|.
\end{equation}
Therefore, the function $f(\tilde{W}) := \left\Vert (GT-In) - \left( \tilde{W} (W^{T}(GT-In)) \right) \right\Vert^2$ is $L$-smooth.
$\hfill\blacksquare$\\

\noindent \textbf{Lemma 3}. The composite function $g = f \circ \mathcal{R}(\cdot)$ is $L$-smooth.\\

\noindent \textit{Proof. } According to Lemma 2, $f$ is $L$-smooth, so it follows that:
\begin{equation}
\|\nabla g(\tilde{W}_1) - \nabla g(\tilde{W}_2)\| = \|\nabla f(\mathcal{R}(\tilde{W}_1)) - \nabla f(\mathcal{R}(\tilde{W}_2))\| \leq L \|\mathcal{R}(\tilde{W}_1) - \mathcal{R}(\tilde{W}_2)\|.
\label{eq:11}
\end{equation}
Applying the Taylor expansion, we get:
\begin{equation}
\mathcal{R}(\tilde{W}_2) = \mathcal{R}(\tilde{W}_1) + D\mathcal{R}(\tilde{W}_1)(\tilde{W}_2 - \tilde{W}_1) + O(\|\tilde{W}_2 - \tilde{W}_1\|^2).
\label{eq:15}
\end{equation}
According to Lemma 1, $D\mathcal{R}(\tilde{W}_1)$ is the identity mapping, i.e., $D\mathcal{R}(\tilde{W}_1) = I$.
Therefore, the higher-order term $O(\|\tilde{W}_2 - \tilde{W}_1\|^2) = 0$.
Thus, Eq. \eqref{eq:15} can be simplified to:
\begin{equation}
\mathcal{R}(\tilde{W}_2) = \mathcal{R}(\tilde{W}_1) + (\tilde{W}_2 - \tilde{W}_1).
\end{equation}
So,
\begin{equation}
\|\mathcal{R}(\tilde{W}_1) - \mathcal{R}(\tilde{W}_2)\| = \|\tilde{W}_1 - \tilde{W}_2\|.
\label{eq:17}
\end{equation}
Substituting Eq. \eqref{eq:17} into Eq. \eqref{eq:11}, we have:
\begin{equation}
\|\nabla g(\tilde{W}_1) - \nabla g(\tilde{W}_2)\| \leq L \|\mathcal{R}(\tilde{W}_1) - \mathcal{R}(\tilde{W}_2)\|= L \|\tilde{W}_1 - \tilde{W}_2\|.
\end{equation}
Therefore, the composite function $g = f \circ \mathcal{R}$ is $L$-smooth.
$\hfill\blacksquare$

\section*{Appendix C. Experiments}

Tab. \ref{tab:The-results-of} presents the full results of the exposure correction dataset. 
Our analysis, following Afifi et al. \cite{laplacian}, divided the table into three groups.
Group one comprises 3,543 images, depicting well- and over-exposed images captured with 0, +1, and +1.5 EVs.
The second group includes 2,362 under-exposed images with -1 and -1.5 EVs.
All 5,905 images captured with all EVs (-1.5, -1, 0, +1, and +1.5) are in the final group.
LCA obtains the highest PSNR in 17 cases and the highest SSIM in 12 cases, which is the best performance among 11 methods.
The second-best method (proposed by Eyiokur et al. \cite{baseline}) obtains the highest PSNR in 4 cases and the highest SSIM in 11 cases.
Although Deep UPE, RetinexNet, and SCI are designed specifically for correcting under-exposure values, LCA still performs better on average than these methods in under-exposure cases.

\begin{table*}[!h]
\begin{centering}
{\footnotesize{}\caption{The results of the exposure dataset.
The best results are highlighted with \textcolor{red}{red}, and the second-best results with \textcolor{blue}{blue}.
The table is divided into three subgroups: 1) 3543 well- and over-exposure images, 2) 2362 under-exposure images, 3) all 5905 images.
Methods with {*} were designed for only the under-exposure case, so they are only presented in the second group. 
}
\label{tab:The-results-of}
}{\footnotesize\par}
\par\end{centering}
\centering{}\resizebox{1.00\textwidth}{!}{
\begin{tabular}{@{}l@{}lccccccccccccc}
\toprule 
\multirow{2}{*}{Group} & \multirow{2}{*}{\;Method} & \multicolumn{2}{c}{Expert A} & \multicolumn{2}{c}{Expert B} & \multicolumn{2}{c}{Expert C} & \multicolumn{2}{c}{Expert D} & \multicolumn{2}{c}{Expert E} & \multicolumn{2}{c}{Avg} & \multirow{2}{*}{PI}\tabularnewline
\cmidrule{3-14} 
    &  & PSNR & SSIM & PSNR & SSIM & PSNR & SSIM & PSNR & SSIM & PSNR & SSIM & PSNR & SSIM
\tabularnewline
\midrule
    & DPE \cite{DPE} & 14.79 & 0.64 & 15.52 & 0.65 & 15.63 & 0.67 & 16.59 & 0.66 & 17.66 & 0.68 & 16.04 & 0.66 & 2.62
\tabularnewline
    & Zero-DCE \cite{Zero-DCE} & 10.12 & 0.50 & 10.77 & 0.50 & 10.40 & 0.51 & 11.47 & 0.52 & 12.35 & 0.56 & 11.02 & 0.52 & 2.77
\tabularnewline
Well- and  & Afifi et al. \cite{laplacian} & 18.87 & 0.74 & 19.57 & 0.72 & 19.79 & 0.76 & 18.82 & 0.71 & 18.94 & 0.72 & 19.20 & 0.73 & 2.18
\tabularnewline
over-exposure & PEC \cite{PEC} & 14.53 & 0.66 & 14.96 & 0.71 & 14.37 & 0.68 & 14.18 & 0.68 & 13.29 & 0.66 & 14.26 & 0.68 & -
\tabularnewline
    & ExReg \cite{multi-dimensional-regressor} & 20.22  & 0.76 & 21.80 & 0.80 & 22.13 & 0.79 & 19.98 & 0.77 & 19.99 & 0.77 & 20.83 & 0.78 & -
\tabularnewline
    & Eyiokur et al. \cite{baseline} & \textcolor{blue}{20.48} & \textcolor{red}{0.86} & \textcolor{blue}{21.83} & \textcolor{blue}{0.89} & \textcolor{blue}{22.44} & \textcolor{red}{0.90} & \textcolor{blue}{20.13} & \textcolor{red}{0.87} & \textcolor{blue}{20.06} & \textcolor{blue}{0.88} & \textcolor{blue}{20.98} & \textcolor{red}{0.88} & \textcolor{blue}{2.16}
\tabularnewline
    & LCA (Ours) & \textcolor{red}{20.81} & \textcolor{red}{0.86} & \textcolor{red}{22.73} & \textcolor{red}{0.90} & \textcolor{red}{22.62} & \textcolor{blue}{0.89} & \textcolor{red}{20.95} & \textcolor{red}{0.87} & \textcolor{red}{20.91} & \textcolor{red}{0.89} & \textcolor{red}{21.60} & \textcolor{red}{0.88} & \textcolor{red}{1.90}
\tabularnewline
\midrule
\multirow{10}{*}{Under-exposure} & RetinexNet{*} \cite{RetinexNet} & 11.68 & 0.61 & 12.71 & 0.61 & 12.13 & 0.62 & 12.72 & 0.62 & 13.23 & 0.64 & 12.49 & 0.62 & 2.93
\tabularnewline
    & DPE \cite{DPE} & 20.15 & 0.74 & 20.97 & 0.70 & 20.92 & 0.74 & 19.05 & 0.69 & 17.51 & 0.65 & 19.72 & 0.70 & 2.56
\tabularnewline
    & Deep UPE{*} \cite{Deep_UPE} & 17.83 & 0.73 & 19.06 & 0.75 & 18.76 & 0.75 & 19.64 & 0.74 & 20.24 & 0.74 & 19.11 & 0.74 & 2.43
\tabularnewline
    & Zero-DCE \cite{Zero-DCE} & 13.94 & 0.59 & 15.24 & 0.59 & 14.55 & 0.59 & 15.20 & 0.59 & 15.89 & 0.61 & 14.96 & 0.59 & 3.00
\tabularnewline
    & Afifi et al. \cite{laplacian} & 19.48 & 0.75 & 20.55 & 0.73 & 20.52 & 0.77 & 18.94 & 0.72 & 18.76 & 0.72 & 19.65 & 0.74 & \textcolor{blue}{2.34}
\tabularnewline
    & SCI{*} \cite{SCI} & 12.72 & 0.67 & 13.68 & 0.74 & 13.21 & 0.71 & 14.50 & 0.74 & 15.58 & 0.78 & 13.94 & 0.73 & -
\tabularnewline
    & PEC \cite{PEC} & 10.12 & 0.57 & 10.75 & 0.64 & 10.46 & 0.61 & 11.50 & 0.64 & 12.55 & 0.68 & 11.08 & 0.63 & -
\tabularnewline
    & ExReg \cite{multi-dimensional-regressor} & 20.16 & 0.75 & 21.71 & 0.79 & 21.94 & 0.78 & 19.86 & 0.76 & 19.99 & 0.76 & 20.73 & 0.77 & -
\tabularnewline
    & Eyiokur et al. \cite{baseline} & \textcolor{blue}{20.40} & \textcolor{blue}{0.86} & \textcolor{blue}{21.68} & \textcolor{blue}{0.88} & \textcolor{red}{22.18} & \textcolor{red}{0.89} & \textcolor{blue}{19.77} & \textcolor{red}{0.87} & \textcolor{red}{19.51} & \textcolor{red}{0.87} & \textcolor{blue}{20.71} & \textcolor{red}{0.87} & 2.38
\tabularnewline
    & LCA (Ours) & \textcolor{red}{20.87} & \textcolor{red}{0.86} & \textcolor{red}{22.69} & \textcolor{red}{0.90} & \textcolor{blue}{22.09} & \textcolor{blue}{0.87} & \textcolor{red}{19.95} & \textcolor{blue}{0.86} & \textcolor{blue}{19.06} & \textcolor{blue}{0.86} & \textcolor{red}{20.93} & \textcolor{red}{0.87} & \textcolor{red}{2.01}
\tabularnewline
\midrule
\multirow{7}{*}{All} & DPE \cite{DPE} & 16.93 & 0.68 & 17.70 & 0.67 & 17.74 & 0.70 & 17.57 & 0.67 & 17.60 & 0.67 & 17.51 & 0.68 & 2.62
\tabularnewline
    & Zero-DCE \cite{Zero-DCE} & 11.64 & 0.54 & 12.56 & 0.54 & 12.06 & 0.54 & 12.96 & 0.55 & 13.77 & 0.58 & 12.60 & 0.55 & 2.87
\tabularnewline
    & Afifi et al. \cite{laplacian} & 19.11 & 0.74 & 19.96 & 0.72 & 20.08 & 0.76 & 18.87 & 0.71 & 18.86 & 0.72 & 19.38 & 0.73 & 2.25
\tabularnewline
    & PEC \cite{PEC} & 12.76 & 0.63 & 13.28 & 0.68 & 12.80 & 0.65 & 13.10 & 0.67 & 12.99 & 0.67 & 12.99 & 0.64 & -
\tabularnewline
    & ExReg \cite{multi-dimensional-regressor} & 20.20 & 0.76 & 21.77 & 0.80 & 22.05 & 0.79 & 19.93 & 0.77 & 19.99 & 0.77 & 20.79 & 0.78 & -
\tabularnewline
    & Eyiokur et al. \cite{baseline} & \textcolor{blue}{20.44} & \textcolor{blue}{0.86} & \textcolor{blue}{21.773} & \textcolor{blue}{0.89} & \textcolor{blue}{22.33} & \textcolor{red}{0.90} & \textcolor{blue}{19.98} & \textcolor{blue}{0.87} & \textcolor{blue}{19.84} & \textcolor{red}{0.88} & \textcolor{blue}{20.87} & \textcolor{red}{0.88} & \textcolor{blue}{2.24}
\tabularnewline
    & LCA (Ours) & \textcolor{red}{20.83} & \textcolor{red}{0.86} & \textcolor{red}{22.71} & \textcolor{red}{0.90} & \textcolor{red}{22.41} & \textcolor{blue}{0.88} & \textcolor{red}{20.55} & \textcolor{red}{0.87} & \textcolor{red}{20.17} & \textcolor{blue}{0.87} & \textcolor{red}{21.33} & \textcolor{red}{0.88} & \textcolor{red}{1.94}
\tabularnewline
\bottomrule
\end{tabular}}
\vspace{-0.3cm}
\end{table*}

\newpage
\begin{figure}[!h]
\begin{centering}
\includegraphics[width=1\columnwidth]{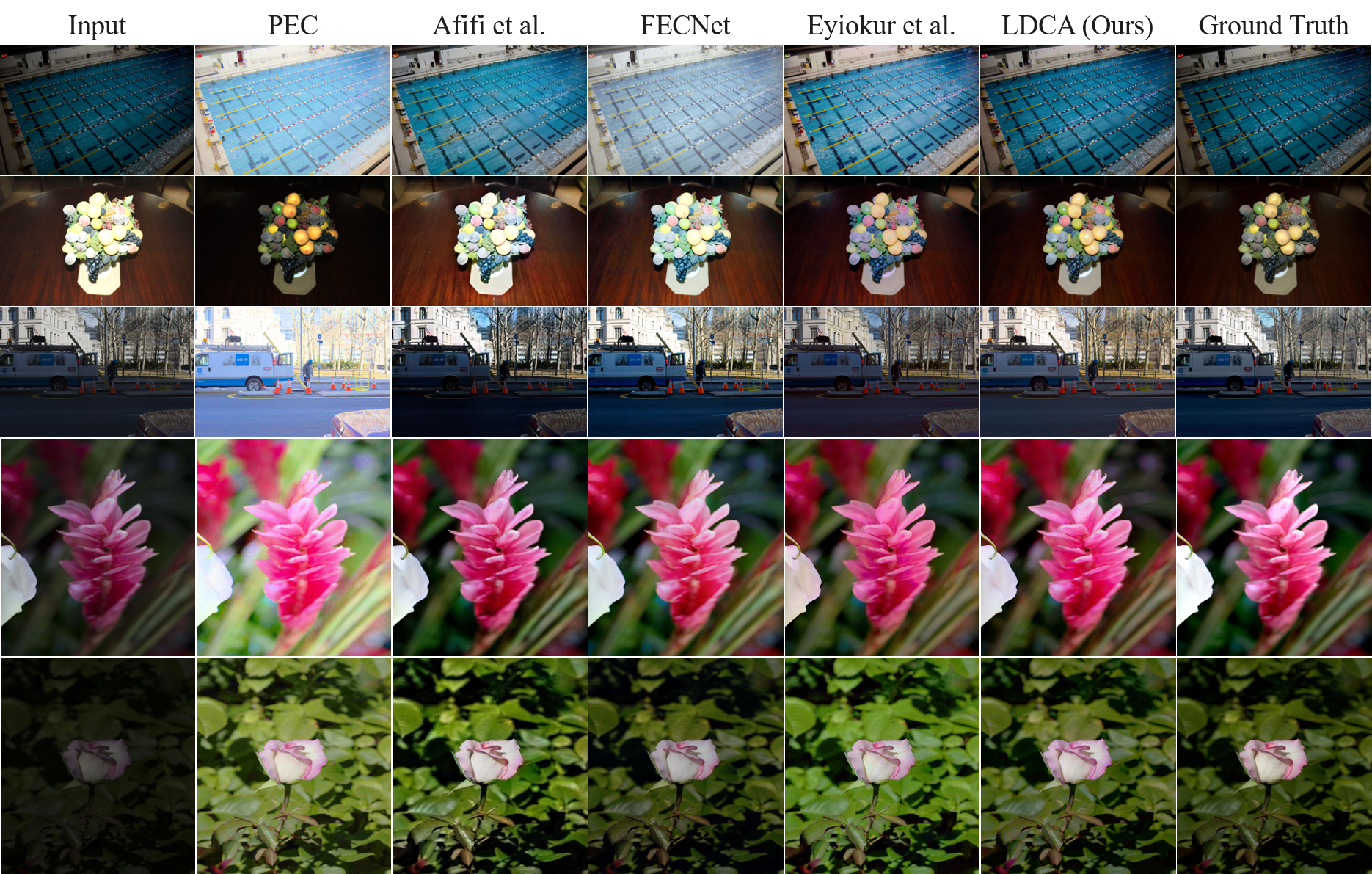}
\par\end{centering}
\caption{Visualization results on the exposure correction dataset.}
\label{fig:Visualization-results-on}
\vspace{-0.2cm} 
\end{figure}

To visually show the exposure correction performance of LCA, we take five cases for comparison.
Fig. \ref{fig:Visualization-results-on} shows the correction results of five methods and input and ground truth images.
The output images from LCA are the closest to the ground truth images in under-exposure or over-exposure cases.

\begin{figure}[!h]
\begin{centering}
\includegraphics[width=0.65\columnwidth]{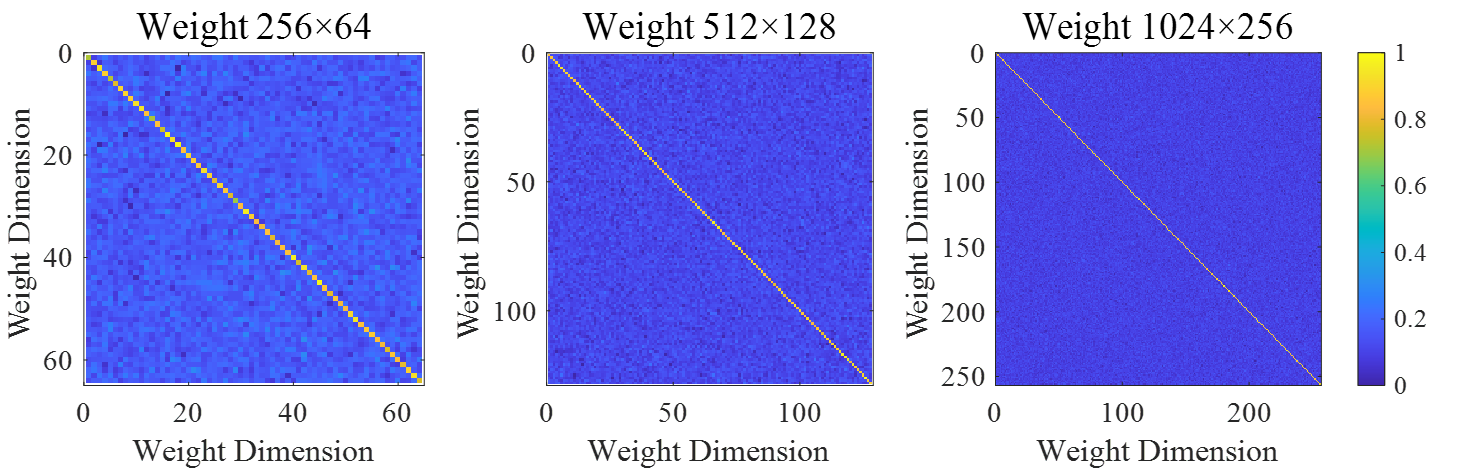}
\par\end{centering}
\caption{Orthogonality visualization of the weights of OLD.}
\label{fig:Orthogonality-visualization-of-th}
\vspace{-10cm} 
\end{figure}

In addition, to show the orthogonality of the OLD optimized by the geometric optimization on Manifolds, we plot weights ($\tilde{W}^{T}\tilde{W}$) from the last three OLDs, as can be seen in Fig. \ref{fig:Orthogonality-visualization-of-th},
which can be found that the weights $\tilde{W}$ of OLD have the orthogonal constraint.
Besides, we also optimize the constrained problem through the barrier penalty function and compare its performance with that optimized by the geometric optimization on Manifolds, which can be seen in Tab. \ref{tab:Comparison-of-the-geo}.
The performance optimized by the barrier penalty function is lower than our methods.
This demonstrates that taking the constraint problem as a \textquotedblleft black box\textquotedblright{} while ignoring the underlying orthogonal relationship could reduce performance. 

\begin{table}[h]
\caption{Comparison of the geometric optimization on Manifolds and barrier penalty function.}
\label{tab:Comparison-of-the-geo}
\centering{}%
\begin{tabular}{lcc}
\hline 
\multirow{2}{*}{Optimisation Method} & \multirow{2}{*}{PSNR} & \multirow{2}{*}{SSIM}\tabularnewline
 &  & \tabularnewline
\hline 
Barrier penalty function & 20.43 & 0.85\tabularnewline
Geometric optimization on Manifolds (Ours)& 20.83 & 0.86\tabularnewline
\hline 
\end{tabular}
\vspace{-0.45cm} 
\end{table}

\end{document}